\documentclass[runningheads]{llncs}

\usepackage{accv}

\usepackage{accvabbrv}

\usepackage{graphicx}
\usepackage{booktabs}
\usepackage{wrapfig}
\usepackage{amssymb}
\usepackage{pifont}
\usepackage{bbm}
\usepackage{colortbl}
\usepackage{amsmath}
\usepackage{mathtools}

\usepackage{caption,subcaption}

\usepackage{multirow}
\usepackage[dvipsnames]{xcolor}
\definecolor{LightRed}{rgb}{0.96,0.92,0.92}
\definecolor{LightGrey}{rgb}{0.9,0.9,0.9}
\definecolor{applegreen}{rgb}{0.0, 0.5, 0.0}

\definecolor{saptarshi}{rgb}{0.54, 0.81, 0.94}
\definecolor{alex}{rgb}{0.19, 0.55, 0.30}
\definecolor{dima}{rgb}{1, 0.2, 0.2}

\usepackage{dsfont}
\usepackage[T1]{fontenc}

\usepackage[breaklinks,colorlinks,citecolor=accvblue]{hyperref}

\usepackage{orcidlink}
\usepackage{arydshln}

\newcommand\tstrut{\rule{0pt}{2.4ex}}
\newcommand\bstrut{\rule[-1.0ex]{0pt}{0pt}}

\begin{document}

\title{Every Shot Counts: Using Exemplars for Repetition Counting in Videos}

\titlerunning{Every Shot Counts}

\author{\href{https://sinhasaptarshi.github.io/}{\textcolor{black}{Saptarshi Sinha}}\inst{1} \and
\href{https://alexandrosstergiou.github.io/}{\textcolor{black}{Alexandros Stergiou}}\inst{2} \and
\href{https://dimadamen.github.io/}{\textcolor{black}{Dima Damen}}\inst{1}}

\authorrunning{S. Sinha et al.}

\institute{$^1$University of Bristol, UK $\quad ^2$University of Twente, NL \\
\href{https://sinhasaptarshi.github.io/escounts}{https://sinhasaptarshi.github.io/escounts}}

\maketitle

\begin{abstract}

Video repetition counting infers the number of repetitions of recurring actions or motion within a video. We propose an exemplar-based approach that discovers visual correspondence of video exemplars across repetitions within target videos. Our proposed \textbf{E}very \textbf{S}hot \textbf{Counts} (ESCounts) model is an attention-based encoder-decoder that encodes videos of varying lengths alongside exemplars from the same and different videos. In training, ESCounts regresses locations of high correspondence to the exemplars within the video. In tandem, our method learns a latent that encodes representations of general repetitive motions, which we use for exemplar-free, zero-shot inference. Extensive experiments over commonly used datasets (RepCount, Countix, and UCFRep) showcase ESCounts obtaining state-of-the-art performance across all three datasets. 
Detailed ablations further demonstrate the effectiveness of our method.

\end{abstract}

\section{Introduction}
In recent years, tremendous progress has been made in video understanding. Visual Language Models (VLMs) have been adopted for many vision tasks including video summarisation~\cite{pramanick2023egovlpv2,lin2023univtg,wu2022intentvizor}, localisation~\cite{ramakrishnan2023naq,wang2022internvideo}, and question answering (VQA)~\cite{alayrac2022flamingo,huang2023clover,mangalam2024egoschema,ye2023hitea}. Despite their great success, recent analysis~\cite{jiang2023clip} shows that VLMs can still fail to count objects or actions correctly. 
Robust counting can be challenging due to appearance diversity, limited training data, and the semantic ambiguity of identifying `what' to count.

Evidence in developmental psychology and cognitive neuroscience~\cite{slaughter2011learning,wang2019infants,wang2023aspects} shows that infants fail to differentiate the number of hidden objects if not shown and counted to them first, suggesting an upper limit of individual objects in working memory. However, infants exposed to an instance of the object first could better approximate carnality. This shows that counting is a visual exercise of matching to exemplars, and is developed before understanding their semantics.

Object-counting in images has recently exploited exemplars to improve performance~\cite{lu2019class,countr}. In training, models attend to one or more exemplars of `what' object(s) to count alongside learnt embeddings for exemplar-free counting.
During inference, only the learnt embeddings are used for zero-shot counting without knowledge from exemplars.
As videos are of variable length and repetition durations vary, these approaches are not directly applicable to videos.

\begin{figure}[tb]
\centering
\includegraphics[width=1\linewidth,trim={0cm 30cm 12cm 0cm},clip]{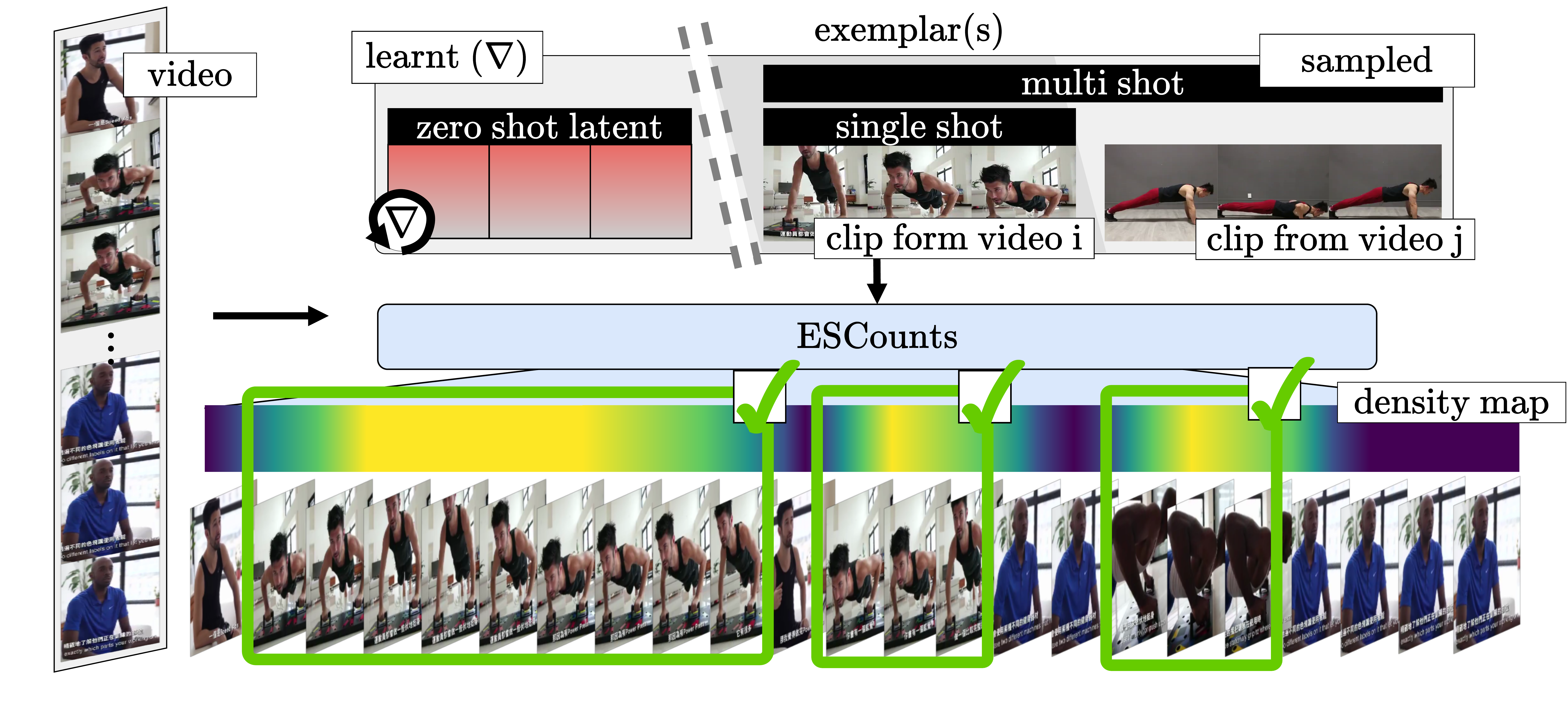}
\vspace{-0.5em}
\caption{\textbf{VRC with ESCounts} involves exemplars for relating information of the repeating action across the video. We visualise the density map with high relevance regions to the action \emph{push-up} being highlighted, whilst regions of low relevance are not.}
\vspace{-1.5em}
\label{fig:teaser}
\end{figure}

Taking inspiration from image-based approaches, we address Video Repetition Counting (VRC) with exemplars for the first time. 
We differ from prior works that formulate VRC as classifying a preset number of repetitions~\cite{bacharidis2023repetition,dwibedi2020counting,zhang2021repetitive}, or detecting relevant parts (start/end) of repetitions~\cite{destro2024cyclecl,hu2022tranrac,li2024repetitive}. Instead, we argue that learning correspondences to reference exemplar(s) during training can provide a strong prior for discovering correspondences across repetitions at inference. We propose \textbf{E}very-\textbf{S}hot \textbf{Counts} (ESCounts), a transformer-based encoder-decoder that during training encodes videos of varying lengths alongside exemplars and learns latents of general repeating motions, as shown in~\cref{fig:teaser}. Similar to~\cite {hu2022tranrac,li2024repetitive}, we use density maps to regress the temporal location of repetitions. At inference, learnt latents are used for exemplar-free counting.

In summary, our contributions are as follows: (i) We introduce exemplar-based counting for VRC (ii) We propose an attention-based encoder-decoder that corresponds exemplars to a query video of varying length. (iii) We learn latents for general repeating features and use them to predict the number of repetitions during inference without exemplars, (iv) We evaluate our approach on the three commonly-used VRC datasets: RepNet~\cite{hu2022tranrac}, Countix~\cite{dwibedi2020counting}, and UCFRep~\cite{zhang2020context}.
Our approach achieves a new state-of-the-art in every benchmark, even on Countix where start-end times of repetitions are not annotated.

\section{Related Works}

We first review methods for the long-established task of object counting in images. We then review VRC methods for videos.

\subsection{Object Counting in Images}

Methods can be divided into class-specific and class-agnostic object counting. 

\noindent \textbf{Class-specific counting}. These methods learn to count objects of singular classes or sets of categories~\eg people~\cite{wang2009crowdhead,li2008estimatingheadshoulders}, cars~\cite{droneobjectcounting}, or wildlife~\cite{countinginthewild}. 
A large portion of object-counting approaches~\cite{droneobjectcounting,wang2009crowdhead,countingeverydayobjects,noroozi2017representation,segui2015learning} have relied on detecting target objects and counting their instances. Traditional methods have used hand-crafted feature descriptors to detect human heads~\cite{wang2009crowdhead} or head-shoulders~\cite{li2008estimatingheadshoulders} for crowd-counting. 
Other methods have used blobs~\cite{detection_based_counting}, individual points~\cite{liu2019recurrent}, and object masks~\cite{cholakkal2019object} for detecting and counting instances. Though object detection can be a preliminary step before counting, detection methods rely strongly on the object detector's performance which can be less effective in densely crowded images~\cite{countingeverydayobjects}. Other methods instead relied on regression, to either regress to the target count~\cite{countingeverydayobjects,xiong2019open} or estimate a density map~\cite{lempitsky2010learning,perspectivefreecounting,zhang2016single}. 

\noindent \textbf{Class-agnostic counting}. Class-specific counting approaches are impractical for general settings where prior knowledge of the object category is not available. Recent works~\cite{countr,countx,oneshotobjectcounting,bmn-net} have used one (or a few) exemplars as references to estimate a density map for unknown target classes.
Building on the property of \textit{image self-similarity},~\cite{lu2019class} proposed a convolutional matching network. They cast counting as an image-matching problem, where exemplar patches from the same image are used to match against other patches within the image. Following up, Liu~\etal~\cite{countr} used an encoder for the query image, a convolution-based encoder for the exemplar, and an interaction module to cross-attend information between the exemplar and the image. A convolutional decoder was used to regress the density map. Recent approaches have also fused text and visual embeddings~\cite{countx}, used contrastive learning across modalities~\cite{jiang2023clip}, and generated exemplar prototypes using stable diffusion~\cite{xu2023zeroshotlvm}. 
Inspired by these methods, we propose an attention-based encoder-decoder that extends exemplar-based counting to VRC. Our approach is invariant to video lengths and can use both learnt or encoded exemplars. 

\subsection{Video Repetition Counting (VRC)}

\looseness-1 Compared to image-based counting, video repetition counting has been less explored.
Early approaches have compressed motion into a one-dimensional signal and recovered the repetition structure from the signal's period~\cite{fourier5_generic,fourier6_periodic,lu2004repetitive,panagiotakis2018unsupervised}. The periodicity can then be counted with Fourier analysis~\cite{fourier1,fourier2_extraction,fourier3_robust,fourier4_visual,fourier5_generic,fourier6_periodic}, peak detection~\cite{thangali2005periodic}, or wavelet analysis~\cite{runia2018real}. However, these methods are limited to uniformly periodic repetitions. For non-periodic repetitions, temporal understanding frameworks~\cite{TAL_chao,TAL_long,TAL_shou,lin2018bsn} have been adapted. Zhang~\etal~\cite{zhang2020context} proposed a context-aware scale-insensitive framework to count repetitions of varying scales and duration. Their method exhaustively searches for pairs of consecutive repetitions followed by a prediction refinement module. 
Recent methods~\cite{destro2024cyclecl,dwibedi2020counting,hu2022tranrac} have also extended image self-similarity to the temporal dimension with Temporal Self-similarity Matrices (TSM). TSM is constructed using pair-wise similarity of embeddings over temporal locations. RepNet~\cite{dwibedi2020counting} used a transformer-based period predictor. To count repetitions with varying speeds, Trans-RAC~\cite{hu2022tranrac} modified TSM to use multi-scale sequence embeddings. For counting under poor lighting conditions,~\cite{zhang2021repetitive} used both audio and video in a multi-modal framework. They selectively aggregated information from the two modalities using a reliability estimation module. 
Li~\etal~\cite{li2024repetitive} also used multi-modal inputs with optical flow as an additional signal supporting RGB for detecting periodicity.

Recent works attempt to utilise spatial~\cite{yao2023poserac} or temporal~\cite{zhao2024skim,li2024efficient} saliency for repetition counting. 
Yao~\etal~\cite{yao2023poserac} proposed a lightweight pose-based transformer model that used action-specific salient poses as anchors. The need for salient pose labels for each action limits generalisability to unseen repetitions.
Zhao~\etal~\cite{zhao2024skim} used a dual-branch architecture to first select repetition-relevant video segments and then attend over these frames. Li~\etal~\cite{li2024efficient} used a joint objective to localise and binary classify regions as (non-)repetitive.

The above methods do not utilise the correspondences discovered by exemplar repetitions.
Thus, do not relate variations in the action's performance.
We propose using action exemplars as references for VRC. Exemplars have previously been used in videos for action recognition tasks~\cite{gaidon2013temporal,jain2020actionbytes,willems2009exemplar,weinland2008action}.~\cite{weinland2008action} used silhouette/pose exemplars for classifying action sequences into predefined categories.\cite{willems2009exemplar} converted training videos to a visual vocabulary and used the most discriminative visual words as exemplars. These methods are limited to a predefined set of classes. To our knowledge, we are the first to use exemplars for repetition counting in videos.

\section{Every Shot Counts (ESCounts) Model}
\label{sec:method}

\begin{figure}[tb]
  \centering
  \includegraphics[width=\linewidth]{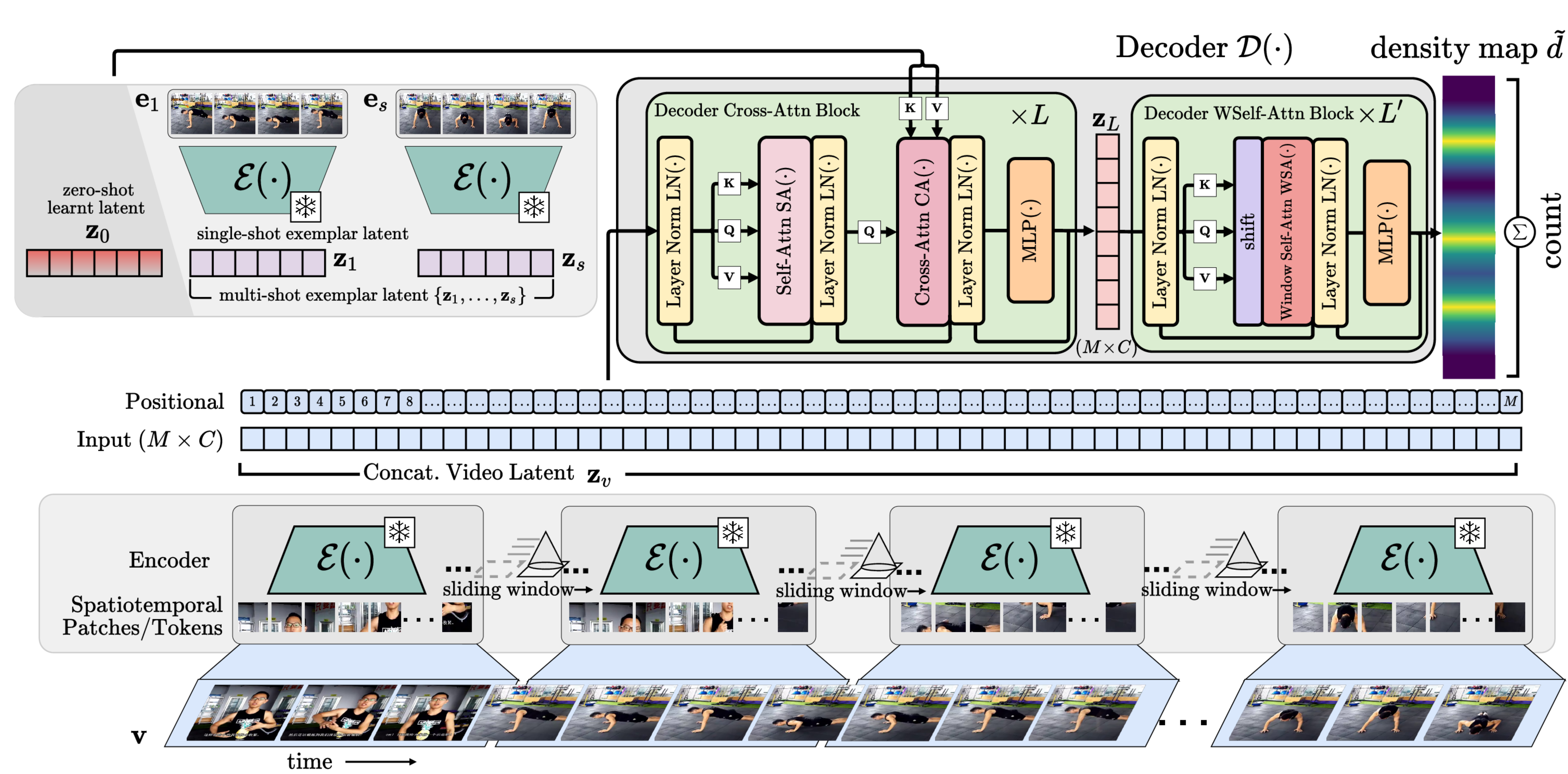}
  \caption{\textbf{ESCounts Model overview}. \textbf{Bottom:} Video $\mathbf{v}$ is encoded by $\mathcal{E}$ over sliding temporal windows to spatiotemporal latents $\mathbf{z}_{v} \in \mathbb{R}^{M \times C}$. \textbf{Top Left:} Exemplars $\{\mathbf{e}_{s}\}$ are also encoded with $\mathcal{E}$. \textbf{Top Right:} Video $\mathbf{z}_{v}$ and exemplar $\mathbf{z}_{s}$ latents are cross-attended by decoder $\mathcal{D}$ over $L$ cross-attention blocks. The resulting $\mathbf{z}_L \in \mathbb{R}^{M \times C}$ are attended over $L'$ window self-attention blocks and projected into density map $\tilde{\mathbf{d}}$. The decoder $\mathcal{D}$ is trained to regress the error between predicted $\tilde{\mathbf{d}}$ and ground truth $\mathbf{d}$ density maps. At inference, the count is obtained by summing $\tilde{\mathbf{d}}$.}
  \label{fig:arch}
  \vspace{-2.0em}
\end{figure}

In this section, we introduce our ESCounts model (overviewed in~\cref{fig:arch}).
We formally define encoding variable length videos alongside our model's output, in~\cref{sec:method::def}. We introduce the attention-based decoder that corresponds the input video to training exemplars and learnt latents
in~\cref{sec:method::correspondence}. Predictions over temporally shifted inputs are then combined, detailed in~\cref{sec:method::overlap}.

\subsection{Input Encoding and Output Prediction}
\label{sec:method::def}

We denote the full \textbf{video} as $\mathbf{v}$ of varying $\mathcal{T}$ length and fixed $H \times W$ spatial resolution. Segment $\mathbf{e}_s$ containing a single instance of the repeating action we wish to count, is selected as an \textbf{exemplar}.
Exemplars are defined based on provided $[$start, end$]$ labels of every repetition in the video\footnote{For datasets where the start/end times are not available, pseudo-labels are used instead by uniformly dividing the video by the ground truth count.}.
During training, we select one or more exemplar shots $\mathcal{S} \subseteq \{\mathbf{e}_1,\dots,\mathbf{e}_s\}$. Each training instance is a combination of the query video and the set of exemplars $(\mathbf{v}, \mathcal{S})$.

We tokenise and encode the video $\mathbf{v}$ from its original size $\mathcal{T} \times H \times W$ into spatiotemporal latents $\mathbf{z}_v$. To account for the video's variable length, encoder $\mathcal{E}$ is applied over a fixed-size sliding window. The encoded video is represented by $\mathbf{z}_v \in \mathbb{R}^{M \times C}$ of $M = \mathcal{T'} H' W'$ spatiotemporal resolution with $C$ channels. We note that $M$ is not a fixed number, as it depends on the video's length $\mathcal{T}$. We add sinusoidal positional encoding to account for the relative order of these spatiotemporal latents while accommodating the variable video length.

For training only, we select exemplars $\mathcal{S}$ from either the same video or another video of the same action category; e.g. given a video containing \textit{push-up} actions, we can sample exemplars from other videos showcasing the same action within the training set.
We define a probability $p$ of sampling the exemplar from a different video; i.e. $p=0$ implies exemplars are only sampled from the same video, whereas for $p=1$ exemplars are always sampled from another video\footnote{We ablate $p$ in our experiments.}.
We sample exemplars randomly from the labelled repetitions of the video.
We use $\mathcal{E}$ to encode latent representations from each exemplar $\mathbf{e}_{s} \in \mathcal{S}$. We use the same encoder $\mathcal{E}$ for encoding $\mathbf{v}$ and $\mathbf{e}_{s}$ to enable direct correspondence.

We construct the ground truth \textbf{density map} $\mathbf{d}$ from the labelled repetitions in the video as a 1-dimensional vector. To match the downsampled temporal resolution of our input video $\mathcal{T'}$, we also temporally downsample the ground-truth labels. The density map takes low values ($\approx\! 0$) at temporal locations without repetitions and high values within repetitions. We use a normal distribution $\mathcal{N}$ centred around each repetition with $(\mu_i= \frac{t_s + t_e}{2},\sigma)$, where $t_e$ and $t_s$ are the start and end times of each repetition $i$.
\begin{equation}
    \mathbf{d}_t=
    \sum_i\mathcal{N}(t; \mu_i, \sigma) \quad
  \forall \; t \in \{1,\dots,\mathcal{T}'\}
  \label{eqn:density_maps}
\end{equation}
Note that the sum of the density map $\mathbf{d}$ matches the ground truth count, i.e. $\sum\mathbf{d} = c$ where $c$ is the ground truth count for the video.

\subsection{Latent Exemplar Correspondence}
\label{sec:method::correspondence}

\looseness-1 Given both the encoded video $\mathbf{z}_v = \mathcal{E}(\mathbf{v})$ and exemplars $\mathbf{z}_s = \mathcal{E}(\mathbf{e}_{s}) \; \forall \; \mathbf{e}_s \in \mathcal{S}$, we use an attention-based decoder $\mathcal{D}(\mathbf{z}_v,\mathbf{z}_s)$ to learn a correspondence between every repetition in the video $v$ and the encoded exemplar. 
Decoder $\mathcal{D}$ takes the encoded video $\mathbf{z}_v$ as input and predicts the location of every repetition in the video. 
The decoder outputs a 1-dimensional predicted density map of length $\mathcal{T}'$ corresponding to the occurrences of the repeating action given the exemplars. 

\begin{wrapfigure}{r}{0.44\textwidth}
    \vspace{-2em}
    \centering
    \includegraphics[width=\linewidth]{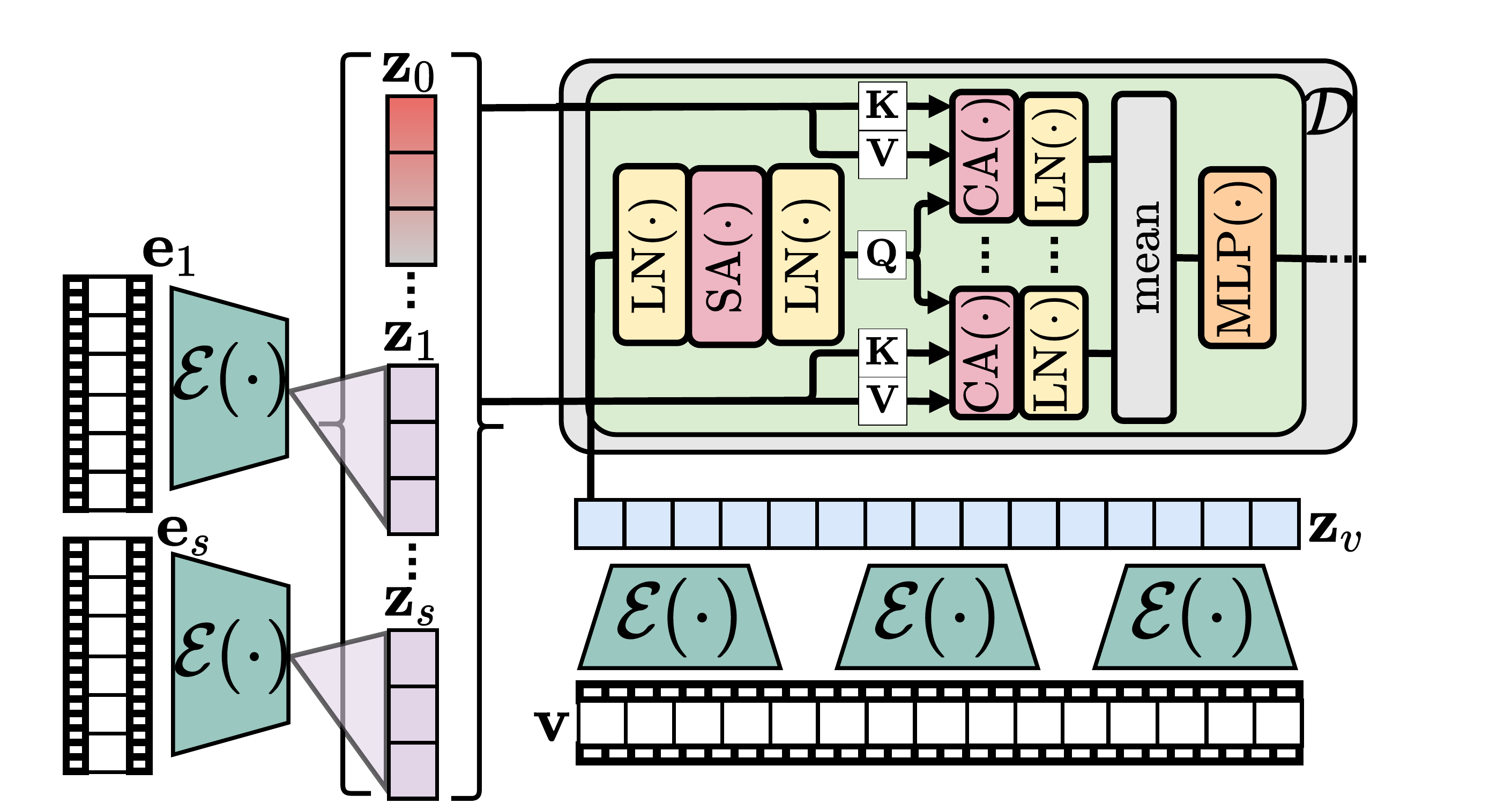}
    \caption{\textbf{Cross-Attention block}. Video latents $\mathbf{z}_v$ are self-attended and then cross-attended with latents $\mathbf{z}_s$ from each exemplar $s \in \mathcal{S}$ and the learnt latent $\mathbf{z}_0$ with the same weights. The resulting representations are then averaged.}
    \label{fig:CAB}
    \vspace{-1em}
\end{wrapfigure}

\noindent
\textbf{Cross-attention Blocks}. We explore the similarity between exemplars and query video representations to predict the corresponding locations of repetitions that match the exemplar. Thus, inspired by~\cite{countr}, we use cross-attention to relate exemplar and video encodings. We define $L$ cross-attention blocks. Each block initially Self-Attends $\text{SA}(\cdot)$ the video latents $\mathbf{z}_l \in \mathbb{R}^{M \times C}$ with multi-head self-attention.  

We note that for the first layer, $\mathbf{z}_1=\mathbf{z}_{v}$. We then relate exemplar and video by Cross-Attending $\text{CA}(\cdot)$ video and exemplar encodings. The block's initial self-attention operation is formulated as:
\begin{equation}
\begin{aligned}
\label{eq:self_att}
    & \mathbf{z}'_{l} = \text{SA}(\text{LN}(\mathbf{z}_{l})) + \mathbf{z}_{l} \, \;\forall \, l \in \,
    \{1,\dots, L\},
\end{aligned}
\end{equation}
\noindent
where $\text{LN}(\cdot)$ is Layer Normalisation. It is essential to self-attend across the video first to capture the features of the repeated actions within the video, and enforce feature correspondence between repetitions.

Repetitions can vary by viewing angles, performance, or duration. We thus wish to allow a varying number of exemplars for counting a repeating action, as shown in~\cref{fig:CAB}. Given a selected number of exemplar shots $\mathcal{S}$, we apply $\text{CA}$ in parallel with $\mathbf{z}'_{l}$ used as a shared query $\mathbf{Q}$ and each of the $\mathcal{S}$ exemplars used as keys and values $\textbf{K},\textbf{V}$ enabling the fusion of repetition-relevant information. As the latents of the video are used as queries $\mathbf{Q}$, spatiotemporal resolution $M$ is maintained. Outputs are then averaged: 
\begin{equation}
\begin{aligned}
\label{eq:cross_att_multishot}
    & \mathbf{z}''_{l} = \frac{1}{N}
    \sum_{s=1}^{\mathcal{S}}\text{CA}(\mathbf{z}_{s},\text{LN}(\mathbf{z}'_{l})) + \mathbf{z}'_l \, \;\forall \, l \in \, 
    \{1,\dots, L\},
\end{aligned}
\end{equation}
where $\mathcal{S}$ is the set of exemplars selected and $\mathbf{z}_{s}$ is the latent for the $s^{\text{th}}$ exemplar. 

\looseness-1 We also want to learn repeating motions to estimate repetitions without explicitly providing exemplars. We thus define a learnable latent $\mathbf{z}_{0}$ to cross-attend $\mathbf{z}_v$. At each training step, we select exemplars from $\{\mathbf{z}_0,\mathbf{z}_1,\text{...},\mathbf{z}_s\}$ and perform $\text{CA}$ with $\mathbf{z}_0$ or $\{\mathbf{z}_1,\text{...},\mathbf{z}_s\}$. \textbf{Importantly, at inference, we use only} $\mathbf{z}_0$.

We obtain the cross-attention blocks' output, defined as $\mathbf{z}_{l+1} \in \mathbb{R}^{M \times C}$, with a Multi-Layer Perceptron MLP on the exemplar-fused latents $\mathbf{z}''_{l}$. 
\begin{equation}
\begin{aligned}
\label{eq:mlp}
    & \mathbf{z}_{l+1} = \text{MLP}(\text{LN}(\mathbf{z}''_{l})) + \mathbf{z}''_{l} \, \;\forall \, l \in \, 
    \{1,\dots, L\}
\end{aligned}
\end{equation}

\noindent
\textbf{Window Self-attention Blocks}. We explore the spatio-temporal inductive bias within the self-attention blocks. For this,
each latent attends locally to its spatio-temporal neighbouring tokens, over 
$L'$
Window Self-Attention $\text{WSA}(\cdot)$~\cite{liu2022video} layers.
We denote $\forall \, l \in \,    \{L+1,\dots,L+L'\}$ :
\begin{equation}
\label{eq:self_att_block}
    \mathbf{z}_{l+1} = \text{MLP}(\text{LN}(\mathbf{z}'_{l})) + \mathbf{z}'_{l},\, \text{where} \; \mathbf{z}'_{l} =
    \begin{cases}
        \text{WSA}(\text{LN}(\mathbf{z}_{l})) + \mathbf{z}_{l}, \; \text{if} \; l=L+1\\
        \text{WSA}(\text{shift}(\text{LN}(\mathbf{z}_{l}))) + \mathbf{z}_{l}, \; \text{else}
    \end{cases}
\end{equation}
where $\text{WSA}$ is window self-attention. Note that following~\cite{liu2022video} windows are shifted at each layer to account for connections across different windows.

The output of the WSA blocks is of size $\mathbf{z}_{L+L'} \in \mathbb{R}^{M \times C}$
. In turn, $\mathbf{z}_{L+L'}$ encodes repetition-relevant features over space and time and is used to predict density map $\tilde{\mathbf{d}}$ for the occurrences of the target repeating action over time. 
We use a fully connected layer to project the latent to a 1-channel vector, i.e. MLP:~${\mathbb{R}^{M\times C} \rightarrow \mathbb{R}^{M}}$.
We then vectorise the spatial resolution $H'W'$ whilst maintaining $\mathcal{T}'$ resulting to the predicted density map $\tilde{\mathbf{d}} \in \mathbb{R}^{\mathcal{T}'}$.

\noindent
\textbf{Training Objective}. Given ground-truth $\mathbf{d}$ and the predicted $\tilde{\mathbf{d}} = \mathcal{D}(\mathbf{z}_v,\mathbf{z}_{s})$ density maps, we train $\mathcal{D}$ to regress the \emph{Mean Square Error} between $\mathbf{d}$ and $\tilde{\mathbf{d}}$, and following~\cite{zhang2021repetitive}, the \emph{Mean Absolute Error} between ground truth counts $c$ and the predicted counts $\tilde{c}$ obtained by linearly summing the density map $\tilde{c} = \sum \tilde{\mathbf{d}}$   
\begin{equation}
    \mathcal{L} = \underbrace{ \frac{|| \mathbf{d} - \tilde{\mathbf{d}}|| ^2}{\mathcal{T}'}}_{\text{MSE}(\mathbf{d},\tilde{\mathbf{d}})} +  \underbrace{\frac{|c - \sum{\tilde{\mathbf{d}}}|}{c}}_{\text{MAE}(c,\tilde{c})} 
\end{equation}

\begin{wrapfigure}{r}{0.4\textwidth}
    \vspace{-1em}
  \begin{center}
    \includegraphics[width=0.38\textwidth]{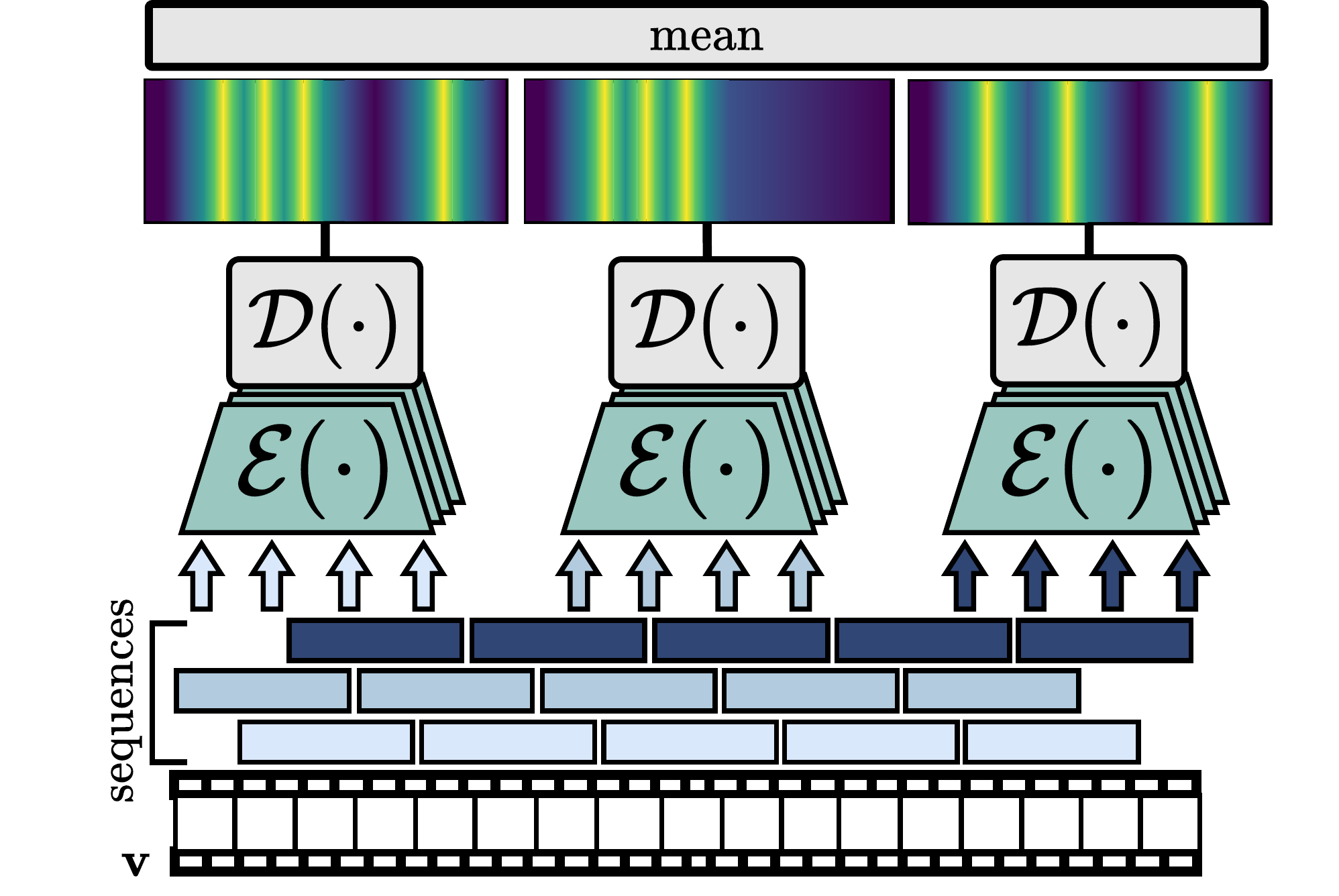}
  \end{center}
  \vspace{-1em}
  \captionof{figure}{\textbf{Shifted Density maps} from each video, are meaned to $\tilde{\mathbf{d}}$.}
  \label{fig:seg}
  \vspace{-4em}
\end{wrapfigure}
\noindent

\noindent
At inference, we use the predicted count $\tilde{c}$.

\subsection{Time-Shift Augmentations}
\label{sec:method::overlap}

The predicted density map $\tilde{\mathbf{d}}$ results from encoding a video with $\mathcal{E}$, over non-overlapping sliding windows. However, as each window is of fixed temporal resolution, repetitions may span over multiple windows. 
Thus, we include time-shifting augmentations in which the start time of the encoded video is adjusted to allow for different spatiotemporal tokens.
We train with augmentations of the start time whilst at inference, we use an ensemble of time-shift augmentations for a more robust estimation. We use multiple overlapping sequences as shown in~\cref{fig:seg} and combine the predicted density maps over $\mathbf{K}$ shifted start/end positions. We obtain the final predicted density map by temporally aligning and averaging the predictions; $\tilde{\mathbf{d}}_t = \frac{1}{|\mathbf{K}|} \underset{k \in \mathbf{K}}{\sum}\tilde{\mathbf{d}}^{k}_{t+\epsilon_k}$, where $\epsilon_k$ is the shifting for each $k \in \mathbf{K}$.

\section{Experiments}
\label{sec:results}

We overview the used datasets, implementation details, and evaluation metrics in~\cref{sec:experimental_setup}. We include quantitative and qualitative comparisons to state-of-the-art methods in~\cref{sec:comparison_to_sota}. We ablate over different ESCounts settings in~\cref{sec:ablation_studies}. 
For all results, we only report zero-shot counting during inference.
In~\cref{sec:inference_with_more_shots}, we evaluate ESCounts' when exemplars are available during inference.

\subsection{Experimental Setup}\label{sec:experimental_setup}
\textbf{Datasets}. We evaluate our method on a diverse set of VRC datasets.

\emph{RepCount}~\cite{hu2022tranrac} contains videos of workout activities with varying repetition durations. Annotations include counts alongside start and end times per repetition. We use the publicly available Part-A with $758$, $131$, and $152$ videos for train, val, and test respectively. Additionally, we use the provided open set split with 70\% categories for train, 10\% for val, and 20\% for testing. We tune the hyperparameters on the \textit{val} set and report our results on the \textit{test} set.

\emph{Countix}~\cite{dwibedi2020counting} is a subset of Kinetics~\cite{carreira2017quo} containing videos of repetitive actions with $4,588$, $1,450$, and $2,719$ videos for train, val, and test respectively. Counts are provided without individual repetition start-end times.
Countix does not have many pauses or interruptions between counts. Thus, we define pseudo-repetition annotations by dividing videos into uniform segments based on the ground truth count. 
The pseudo-labels are used to estimate the density maps without additional annotations, to compare directly to other methods.

\emph{UCFRep}~\cite{zhang2020context} is a subset of UCF-101~\cite{soomro2012ucf101} consisting of 420 train and 106 val videos from 23 categories with counts and annotations of start and end times. Following~\cite{zhang2020context,li2024repetitive}, we report our results on the \textit{val} split as no \textit{test} set is available.

\noindent
\textbf{Implementation Details}. Unless specified otherwise, we use MAE-pretrained ViT-B~\cite{feichtenhofer2022masked} as our encoder $\mathcal{E}$ with Kinetics-400~\cite{carreira2017quo} weights. 
We sample frames from variable-length videos every 4 frames using a sliding window of $64$ frames. At each window, our encoder's input is of $16 \times 224 \times 224$ size, and the output is $8\times14\times14$, resulting in $1568$ spatiotemporal tokens.
We use $C=512$ channels\footnote{when encoders have a different output, we add a fully connected layer to map to $C$}.
The input to the decoder is of variable length $M = 1568\frac{\mathcal{R}}{64}$, where $\mathcal{R}$ is the total number of frames in the video at raw framerate.
Exemplars are sampled uniformly with $16$ frames between the start and end of a repetition. 

\looseness-1 The encoder is frozen and we only train the decoder and zero-shot latent $\mathbf{z}_0$. We use $L=2$ and $L'=3$ ablating this choice in~\cref{sec:ablations2} of the appendix. We train for 300 epochs on a single Tesla V100 with a batch size of 1, to deal with variable-length videos, accumulating gradients over 8 batches. We use $5e^{-2}$ weight decay and a learning rate of $5e^{-5}$ with decay by $0.8$ every $60$ epochs. Per training instance, we randomly set the number of exemplars $| \mathcal{S} | \! \sim \! \{0, 1, 2\}$ and sample $\mathcal{S}$ exemplars. 
We set the chance of sampling exemplars from a different video to $p = 0.4$.

Only the learnt latent are used at inference to predict repetition counts. We aggregate predictions over $|\mathbf{K}|=4$ sequences.

\noindent
\textbf{Evaluation Metrics}. Following previous VRC works~\cite{hu2022tranrac,dwibedi2020counting,zhang2021repetitive}, we use Mean Absolute Error (MAE) and Off-By-One accuracy (OBO) as evaluation metrics, calculated as~\cref{eqn:MAE,eqn:OBO} respectively.  
Inspired by image counting methods~\cite{countr,countx}, we introduce Root-Mean-Square-Error (RMSE) in~\cref{eqn:RMSE} for VRC providing a more robust metric for diverse counts compared to MAE's bias towards small counts.
We also report the off-by-zero accuracy (OBZ) in~\cref{eqn:OBZ} as a tighter metric than the corresponding OBO for precise counts.  

\begin{minipage}{.45\linewidth}
\begin{align}
  MAE = \frac{1}{|\Omega|}\sum_{i \in \Omega} \frac{| c_i - \tilde{c}_i|}{c_i} ,
  \label{eqn:MAE}
\end{align}
\end{minipage}%
\begin{minipage}{.5\linewidth}
\begin{equation}
  OBO = \frac{1}{|\Omega|}\sum_{i \in \Omega} \mathds{1}(| c_i - \tilde{c}_i | \leq 1 ) ,
  \label{eqn:OBO}
\end{equation}
\end{minipage}\newline
\begin{minipage}{.5\linewidth}
\begin{equation}
  RMSE = \sqrt{\frac{1}{|\Omega|}\sum_{i \in \Omega}(  c_i - \tilde{c}_i )^2} ,
\label{eqn:RMSE}
\end{equation}
\end{minipage}%
\begin{minipage}{.5\linewidth}
\begin{equation}
  OBZ = \frac{1}{|\Omega|}\sum_{i \in \Omega} \mathds{1}(| c_i - \tilde{c}_i | = 0) ,
  \label{eqn:OBZ}
\end{equation}
\end{minipage}

\noindent
where  $c_i$, $\tilde{c}_i$ are the ground-truth and predicted counts for $i$-th video in test set $\Omega$. $\mathds{1}$ is the indicator function.

\subsection{Comparison with State-of-the-art}\label{sec:comparison_to_sota}

In \cref{table:sota}, we compare ESCounts, to prior methods on the three datasets.
We provide results on the same backbone as the best-performing method on each dataset, for fair and direct comparison to previous works. 

\begin{table}[tp!]
\caption{\textbf{Comparison of VRC methods}. $\dagger$ represents multi-modal models that use added audio or flow. \textcolor{red}{$^*$} denotes results reproduced using provided checkpoints. \textcolor{darkgray}{$^*$} denotes inhouse re-training using published codes. \textcolor{gray}{Grayed rows} in (c) represent methods that finetune the encoder. Top performances for each metric and dataset are in \textbf{bold}.}
\label{table:sota}
\vspace*{-2em}

\begin{minipage}[t]{0.51\textwidth}
\subcaption{RepCount}
\vspace{-1.5em}
\resizebox{\columnwidth}{!}{
\begin{tabular}[t]{ ll c cccc c cc}
\toprule
& & $\;$ & \multicolumn{4}{c}{benchmark} & $\;$ & \multicolumn{2}{c}{open set} \bstrut \\ \cline{4-7} \cline{9-10}
Method & Encoder && RMSE$\downarrow$ & MAE$\downarrow$ & OBZ$\uparrow$ & OBO$\uparrow$ && MAE$\downarrow$ & OBO$\uparrow$ \tstrut \bstrut \\
\midrule
RepNet~\cite{dwibedi2020counting} & R2D50 && \multicolumn{1}{c}{-} & 0.995 & \multicolumn{1}{c}{-} & 0.013 && \multicolumn{1}{c}{-} & \multicolumn{1}{c}{-}\tstrut \bstrut \\
TransRAC~\cite{hu2022tranrac} & SwinT && $\;$9.130\textcolor{red}{$^*$} & 0.443 & $\;$0.085\textcolor{red}{$^*$} & 0.291 && 0.625 & 0.204\tstrut \bstrut\\
MFL~\cite{li2024repetitive}$\dagger$ & SwinT && \multicolumn{1}{c}{-} & 0.384 & \multicolumn{1}{c}{-} & 0.386 && \multicolumn{1}{c}{-} & \multicolumn{1}{c}{-}\tstrut \bstrut\\
DeTRC~\cite{li2024efficient} & ViT-B && \multicolumn{1}{c}{-} & 0.262 & \multicolumn{1}{c}{-} & 0.543 && \multicolumn{1}{c}{-} & \multicolumn{1}{c}{-}\tstrut \bstrut\\
SkimFocus~\cite{zhao2024skim} & SwinB && \multicolumn{1}{c}{-} & 0.249 & \multicolumn{1}{c}{-} & 0.517 && \multicolumn{1}{c}{-} & \multicolumn{1}{c}{-} \tstrut \bstrut \\
\rowcolor{LightRed}ESCounts & SwinT && 6.905 & 0.298 & $0.183$ & 0.403 && \multicolumn{1}{c}{-} & \multicolumn{1}{c}{-}\tstrut \bstrut \\
\rowcolor{LightRed} ESCounts & ViT-B && $\mathbf{4.455}$ & $\mathbf{0.213}$& $\mathbf{0.245}$ & $\mathbf{0.563}$ && $\mathbf{0.436}$ & $\mathbf{0.519}$\\
\bottomrule
\end{tabular}
}
\label{table:sota_repcount}
\end{minipage}
\hspace{\fill}
\begin{minipage}[t]{0.48\textwidth}
\subcaption{Countix}
\vspace{-1.5em}
\resizebox{\textwidth}{!}{
\begin{tabular}[t]{llcccc}
\toprule
Method & Encoder & RMSE$\downarrow$ & MAE$\downarrow$ & OBZ$\uparrow$ & OBO$\uparrow$ \tstrut \bstrut\\
\midrule
RepNet~\cite{dwibedi2020counting} & R2D50 & - & 0.364 & - &0.697\tstrut \bstrut\\
Sight \& Sound~\cite{zhang2021repetitive}$\dagger$ & R(2+1)D18 & - & 0.307 & - & 0.511 \tstrut \bstrut\\
\rowcolor{LightRed} ESCounts & R(2+1)D18 & 3.536 & 0.293 & 0.286 & $\mathbf{0.701}$ \tstrut \bstrut\\
\rowcolor{LightRed} ESCounts & ViT-B & $\mathbf{3.029}$ & $\mathbf{0.276}$ & $\mathbf{0.319}$ & 0.673 \\
\bottomrule
\end{tabular}}
\label{table:sota_countix}
\end{minipage}
\vspace{-6pt}

\begin{minipage}[t]
{0.44\textwidth}
\vspace{0.3em}
\subcaption{UCFRep}
\vspace{-1.2em}
\resizebox{\textwidth}{!}{
\begin{tabular}[t]{ll llll}
\toprule
Method & Encoder & RMSE$\downarrow$ & MAE$\downarrow$ & OBZ$\uparrow$ & OBO$\uparrow$ \tstrut \bstrut\\
\midrule
Levy \& Wolf~\cite{levy2015live}  & RX3D101 & \multicolumn{1}{c}{-} & 0.286 & \multicolumn{1}{c}{-} & 0.680  \tstrut \bstrut\\
RepNet~\cite{dwibedi2020counting} & R2D50 & \multicolumn{1}{c}{-} & 0.998 & \multicolumn{1}{c}{-} & 0.009 \\
Context (F)~\cite{zhang2020context} & RX3D101 & 5.761\textcolor{darkgray}{$^*$} & 0.653\textcolor{darkgray}{$^*$} & 0.143\textcolor{darkgray}{$^*$} & 0.372\textcolor{darkgray}{$^*$} \\
TransRAC~\cite{hu2022tranrac} & SwinT & \multicolumn{1}{c}{-} & 0.640 & \multicolumn{1}{c}{-} & 0.324 \\
MFL\cite{li2024repetitive}$\dagger$ & RX3D101 & \multicolumn{1}{c}{-} & 0.388 & \multicolumn{1}{c}{-} & 0.510\\
\rowcolor{LightRed} ESCounts & RX3D101 & 2.004 & 0.247 & 0.343 & 0.731 \\
\rowcolor{LightRed} ESCounts & ViT-B & \textbf{1.972} & 0.216 & 0.381 & 0.704 \bstrut \\
\hline
\textcolor{gray}{Context~\cite{zhang2020context}} & \textcolor{gray}{RX3D101} & \textcolor{gray}{2.165}\textcolor{red}{$^*$} & \textcolor{gray}{0.147} & \textcolor{gray}{$\mathbf{0.452}$}\textcolor{red}{$^*$} & \textcolor{gray}{0.790} \tstrut \\
\textcolor{gray}{Sight \& Sound~\cite{zhang2021repetitive}}$\dagger$ & \textcolor{gray}{R(2+1)D18}  &  \multicolumn{1}{c}{\textcolor{gray}{-}} & \textcolor{gray}{$\mathbf{0.143}$} & \multicolumn{1}{c}{\textcolor{gray}{-}} & \textcolor{gray}{$\mathbf{0.800}$} \\
\bottomrule
\end{tabular}
}
\label{table:sota_ucf}
\end{minipage}%
\hfill
\begin{minipage}[t]{0.54\textwidth}
\vspace*{-2pt}
\caption{\footnotesize \textbf{Cross-dataset generalisation}. Arrows denote train $\rightarrow$ test datasets. Results with provided checkpoints are denoted with \textcolor{red}{$^*$}.}
\vspace{-.9em}
\centering
\resizebox{\textwidth}{!}{%
\begin{tabular}{l c cccc c cccc}
\toprule
 & $$ & \multicolumn{4}{c}{RepCount $\rightarrow$ Countix} & $$ & \multicolumn{4}{c}{RepCount $\rightarrow$ UCFRep} \tstrut \bstrut \\ \cline{2-6} \cline{8-11}
 && RMSE$\downarrow$ & MAE$\downarrow$ & OBZ$\uparrow$ & OBO$\uparrow$ & & RMSE$\downarrow$ & MAE$\downarrow$ & OBZ$\uparrow$ & OBO$\uparrow$ \tstrut \bstrut\\ \hline
RepNet \cite{dwibedi2020counting} && - & - & - & - &&  - & 0.998 & - & 0.009 \tstrut \bstrut\\
TransRAC~\cite{hu2022tranrac}&&  6.867\textcolor{red}{$^*$} & 0.593\textcolor{red}{$^*$} & 0.132\textcolor{red}{$^*$} & 0.364\textcolor{red}{$^*$} && 
6.701\textcolor{red}{$^*$} & 0.640 & 0.087\textcolor{red}{$^*$} & 0.324 \\

MFL~\cite{li2024repetitive}&& - & - & - & - && - & 0.523 & - & 0.350 \\
SkimFocus~\cite{zhao2024skim} && - & - & - & - && - & 0.502 & - & 0.391 \\
DeTRC \cite{li2024efficient}&& - & - & - & - && - & 0.543 & - & 0.418 \\
\rowcolor{LightRed} ESCounts &&   $\mathbf{4.429}$ & $\mathbf{0.374}$ & $\mathbf{0.185}$ & $\mathbf{0.521}$ && $\mathbf{3.536}$ & $\mathbf{0.317}$ & $\mathbf{0.219}$ & $\mathbf{0.571}$ \\ \bottomrule
\end{tabular}%
}
\label{table:cross-dataset generalization}
\vspace{-1.5em}
\end{minipage}
\vspace{-1em}
\end{table}

\noindent
\textbf{RepCount}.~\cref{table:sota_repcount} shows that ESCounts outperforms recent methods \cite{hu2022tranrac,li2024repetitive,li2024efficient,zhao2024skim}. 
Compared to the baseline~\cite{hu2022tranrac}, we improve OBZ by $+0.16$ and reduce RMSE by $-4.68$. 
We test on two backbones - SwinT~\cite{liu2022video} used in~\cite{hu2022tranrac,li2024repetitive} and ViT-B used in~\cite{li2024efficient}. 
On the same SwinT backbone, our approach outperforms~\cite{li2024repetitive}, which uses optical flow and video in tandem, by margins of $-0.09$ MAE and $+0.02$ OBO, showcasing ESCounts' ability to learn repeating motions implicitly. 
With a ViT-B backbone, we outperform~\cite{li2024efficient} by $-0.05$ MAE and $+0.02$ OBO.

We additionally compare ESCounts on the open set setting of RepCount-A, with non-overlapping action categories between train and test sets. ESCounts outperforms~\cite{hu2022tranrac} significantly with $-0.19$ MAE and $+0.32$ OBO.
Note that recent works do not report on this more challenging setup.

\noindent
\textbf{Countix}. Compared to the state-of-the-art~\cite{dwibedi2020counting,zhang2021repetitive} in~\cref{table:sota_countix}, our ESCounts consistently outperforms other models with the same R(2+1)D18 encoder. Our video-only model surpasses the audio-visual model in~\cite{zhang2021repetitive} by $+0.19$ OBO.  Further improvements on the RMSE, MAE, and OBZ are observed with ViT-B.

\noindent
\textbf{UCFRep}.
Compared to methods with frozen encoders in~\cref{table:sota_ucf}, ESCounts with ViT-B improves the previous SoTA by $+0.19$ OBO and $-0.17$ MAE and outperforms~\cite{li2024repetitive} on the same RX3D101 backbone by $+0.22$ OBO.
Our method does not outperform~\cite{zhang2021repetitive,zhang2020context} that fine-tune their encoders on UCFRep. As noted in~\cite{li2024repetitive} this is advantageous given the dataset's size. We show this experimentally by reporting Context (F) trained from the available code of~\cite{zhang2020context} with a frozen encoder, resulting in a significant performance drop with +0.51 MAE and -0.42 OBO. In all directly comparable results, ESCounts achieves stronger results.

\begin{figure}[t!]
    \centering
    \includegraphics[width=\linewidth,trim={0 1.5cm 0 0},clip]{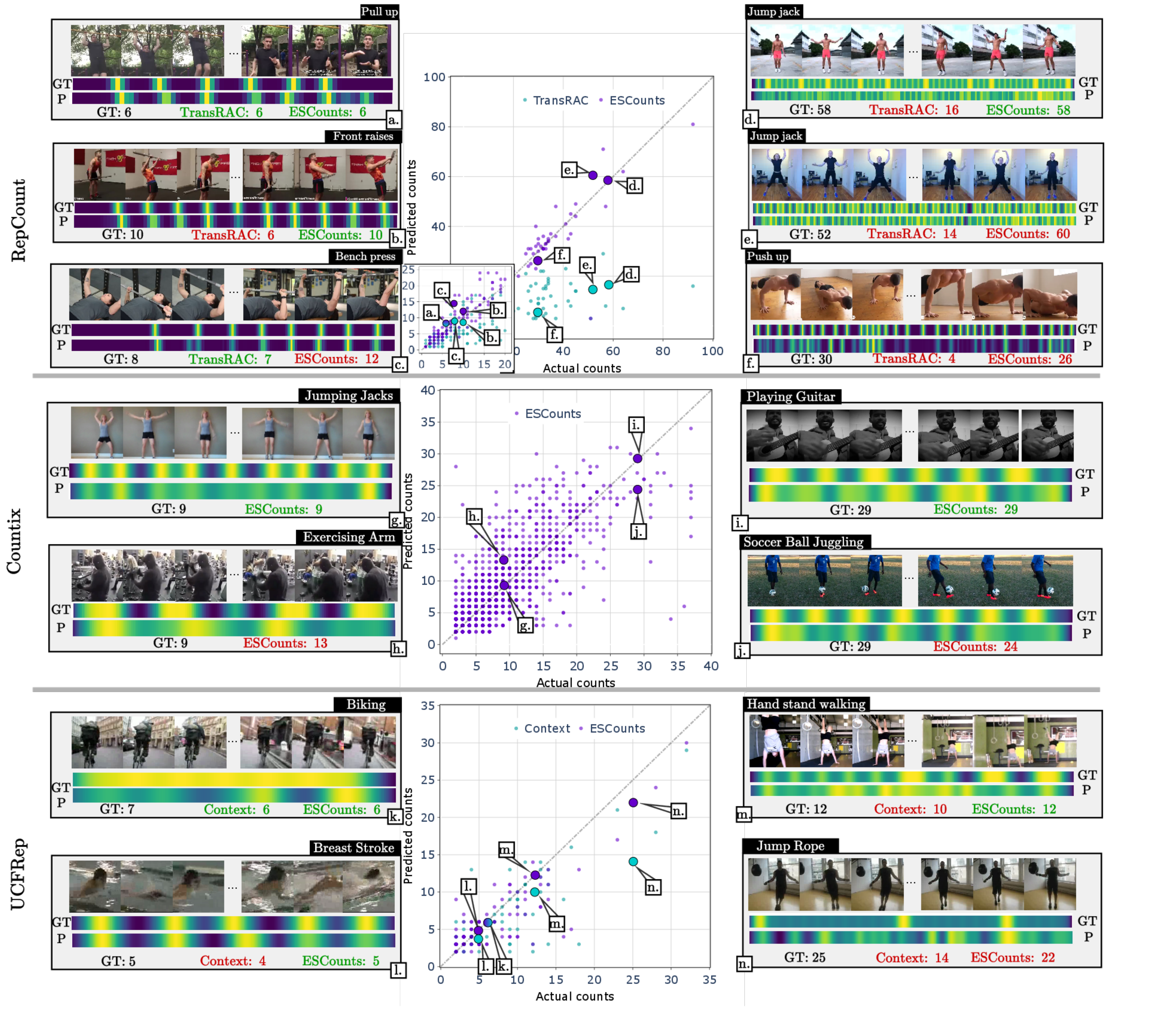}
    \vspace*{-12pt}
    \caption{\textbf{RepCount, Countix, and UCFRep scatter plot, instances, and density maps}. The dotted diagonal denotes correct predictions. We compare ESCounts against TransRAC on Repcount and Context on UCFRep. Action classes and count predictions are shown for each instance. We add the Ground Truth (GT) and Predicted (P) density maps per instance. Pseudo-labels are shown as GT for Countix.
    }
    \vspace{-1.5em}
    \label{fig:qualitative}
\end{figure}

\noindent
\looseness-1 \textbf{Qualitative Results}. In~\cref{fig:qualitative} we visualise predicted to ground truth counts as scatter plots. For RepCount and UCFRep, we select~\cite{hu2022tranrac} and~\cite{zhang2020context} as respective baselines  
and use their publicly available checkpoints\footnote{\cite{li2024repetitive} was not used as a baseline as the code is not public. The publicly available checkpoints on Countix obtained lower results than originally reported.}. 
ESCounts accurately predicts the number of repetitions for a wide range of counts, with most predictions being close to the ground truth i.e. the diagonal. Though predictions from both the baseline and ESCounts are close to the ground truth in low counts, they significantly diverge in high counts. We visualise specific examples and their density maps. ESCounts is robust to the magnitude of counts, with accurate predictions over low (a,b,g,k,l) and high (d,i,m) count examples. In cases of over- and under-predictions; \eg (c,e,f,h,j,n) ESCounts predictions remain closer to actual counts. As shown by the density maps, ESCounts can also localise the repetitions. For Countix, even though ESCounts can predict accurate counts, as the model was trained on pseudo labels, it struggles to localise some of the repetitions. We investigate the localisation capabilities in~\cref{sec:localisation_results} of the appendix.

\noindent
\looseness-1  \textbf{Cross-dataset Generalisation}.
Following~\cite{hu2022tranrac,li2024repetitive}, we test the generalisation capabilities of our method in~\cref{table:cross-dataset generalization}. We use ESCounts trained on RepCount and evaluate on the Countix and UCFRep test sets. For Countix, we outperform the baseline~\cite{hu2022tranrac} by significant margins across metrics. For UCFRep, our method surpasses~\cite{li2024repetitive} by $-0.21$ in MAE and $+0.22$ in OBO. ESCounts in this setting still outperforms~\cite{dwibedi2020counting,hu2022tranrac,li2024repetitive,li2024efficient} \emph{trained} on UCFRep in~\cref{table:sota_ucf}, showcasing the strong ability of ESCounts to generalise to unseen actions.

\subsection{Ablation Studies}\label{sec:ablation_studies}

In this section, we conduct ablation studies on RepCount~\cite{hu2022tranrac} using ViT-B as the encoder.
We study the impact of exemplars by replacing cross- with self-attention and varying the number of training exemplars. We evaluate the sensitivity of our method to the exemplar sampling probability $p$, density map $\sigma$, the impact of time-shift augmentations, and the components of our objective.

\begin{table}[tp]
\caption{\textbf{Ablations on RepCount} over different ESCounts settings.}
\vspace{-1.5em}
\begin{minipage}[t]{0.33\linewidth}
\centering
    \subcaption{SA-only decoder}\label{tab:self-attention_Vs_ours}
    \vspace{-1.5em}
    \resizebox{.93\linewidth}{!}{%
    
   \begin{tabular}[t]{ccccc}
   \toprule
   & RMSE$\downarrow$ & MAE$\downarrow$ & OBZ$\uparrow$ & OBO$\uparrow$ \\
   \midrule
   SA-only & 5.654 & 0.273 & 0.147 & 0.470 \\
  \rowcolor{LightRed}ESCounts & $\mathbf{4.455}$ & $\mathbf{0.213}$  & $\mathbf{0.245}$ & $\mathbf{0.563}$\\
   \bottomrule
   \end{tabular}
    }
\end{minipage}%
\hfill
\begin{minipage}[t]{0.35\linewidth}
\subcaption{Number of exemplars $| \mathcal{S} |$}\label{tab:multishot_in_training}
\centering
\vspace{-1.5em}
\resizebox{\linewidth}{!}{%
    \begin{tabular}[t]{lcccccc}
\toprule
 \textbf{$|\mathcal{S}|$} & \multirow{-1}{*}{RMSE$\downarrow$} & \multirow{-1}{*}{MAE$\downarrow$} & \multirow{-1}{*}{OBZ$\uparrow$} & \multirow{-1}{*}{OBO$\uparrow$} \tstrut \bstrut \\
 \midrule
 $|\mathcal{S}| = 0$ &4.962  & 0.240 & 0.223 & 0.519 \\
 $|\mathcal{S}| \sim \{0,1\}$ & 4.633 & 0.228 & 0.236 &  0.546\\
 $|\mathcal{S}| \sim \{0,2\}$ & 4.601 & 0.226 & 0.239 & 0.550\\
\rowcolor{LightRed} $|\mathcal{S}| \sim\{0, 1, 2\}$ & $\mathbf{4.455}$ & $\mathbf{0.213}$  & 0.245 & $\mathbf{0.563}$ \\
$|\mathcal{S}| \sim \{0,1,2,3\}$ & 4.497 & 0.215 & $\mathbf{0.246}$ & 0.560\\
$|\mathcal{S}| \sim \{0,1,2,3,4\}$ & 4.482 & 0.215 & 0.240 & 0.559\\
\bottomrule
\end{tabular}%
}
\end{minipage}%
\hfill
\begin{minipage}[t]{0.31\linewidth}
\subcaption{Exemplar sampling}\label{tab:class_sampling}
\centering
\vspace{-1.5em}
\resizebox{.95\linewidth}{!}{
\begin{tabular}[t]{llcccc}
\toprule
Diff &same & & & &\\
video &class & \multirow{-2}{*}{RMSE$\downarrow$} & \multirow{-2}{*}{MAE$\downarrow$} & \multirow{-2}{*}{OBZ$\uparrow$} & \multirow{-2}{*}{OBO$\uparrow$} \\ \midrule
\ding{55} &- & 4.701 & 0.224 & 0.226 & 0.521\\
\ding{52} &\ding{55} & 5.553 & 0.270  & 0.165 & 0.464 \\
\rowcolor{LightRed} \ding{52} &\ding{52} & $\mathbf{4.455}$ & $\mathbf{0.213}$  & $\mathbf{0.245}$ & $\mathbf{0.563}$\\

\bottomrule
\end{tabular}
}
\end{minipage}%
\vspace{0.5em}
\begin{minipage}[t]{0.25\linewidth}
\subcaption{Sampling prob. $p$}\label{tab:sampling exemplars}
\centering
\vspace{-1.5em}
\resizebox{.95\linewidth}{!}{
\begin{tabular}[t]{ccccc}
\toprule
$p$ & RMSE$\downarrow$ & MAE$\downarrow$ & OBZ$\uparrow$ & OBO$\uparrow$ \\ \midrule
$0.0$ & 4.919 & 0.240 & 0.205 & 0.545\\
$0.2$ & 4.654 & 0.221 & 0.236& 0.550 \\
\rowcolor{LightRed} $0.4$ & $\mathbf{4.455}$ & $\mathbf{0.213}$  & $\mathbf{0.245}$ & $\mathbf{0.563}$\\
$0.6$ &  4.561 & 0.218  & 0.240 & 0.558 \\
$0.8$ &  4.735& 0.230  & 0.223 & 0.553 \\
$1.0$ & 5.012 & 0.245  & 0.218 & 0.532 \\
\bottomrule
\end{tabular}
}
\end{minipage}%
\hfill
\begin{minipage}[t]{0.25\linewidth}
\subcaption{Density peaks $\sigma$ }\label{tab:varying_width_of_density_peaks}
\centering
\vspace{-.7em}
\resizebox{\linewidth}{!}{%
\begin{tabular}{ ccccc}
\toprule
$\sigma$ &RMSE$\downarrow$ & MAE$\downarrow$ & OBZ$\uparrow$ & OBO$\uparrow$ \\ 
\midrule
Variable & 6.152 & 0.301 & 0.165 & 0.457 \\
0&  5.145 & 0.241 & 0.206& 0.510 \\
0.25& 4.871 & 0.226 & 0.228 & 0.542 \\
\rowcolor{LightRed}0.50& $\mathbf{4.455}$ & $\mathbf{0.213}$ & $\mathbf{0.245}$ & $\mathbf{0.563}$ \\
0.75& 4.683 & 0.218 & 0.240 & 0.556 \\
1.00& 4.732 & 0.223 & 0.238& 0.552\\
\bottomrule
\end{tabular}%
}
\end{minipage}%
\hfill
\begin{minipage}[t]{0.25\linewidth}
\subcaption{Timeshift Aug. $|\mathbf{K}|$}\label{tab:use_overlapping_sequences}
\centering
\vspace{-0.7em}
\resizebox{0.9\linewidth}{!}{%
\begin{tabular}{ ccccc}
\toprule
$|\mathbf{K}|$ &RMSE$\downarrow$ & MAE$\downarrow$ & OBZ$\uparrow$ & OBO$\uparrow$ \\ 
\midrule
1&  4.592 & 0.221 & 0.235& 0.552 \\
2& 4.493 & 0.217  & 0.242 & 0.556\\
3& 4.471 & 0.213 & 0.243 & 0.561 \\
\rowcolor{LightRed}4& $\mathbf{4.455}$ & $\mathbf{0.213}$  & $\mathbf{0.245}$ & $\mathbf{0.563}$\\
\bottomrule
\end{tabular}%
}
\end{minipage}%
\hfill
\begin{minipage}[t]{0.25\linewidth}
\subcaption{Effect of  Objective}\label{tab:effect_of_Ls}
\centering
\vspace{-1.5em}
\resizebox{.95\linewidth}{!}{
\begin{tabular}[t]{lcccc}
\toprule
\multicolumn{1}{c}{Obj} & RMSE$\downarrow$ & MAE$\downarrow$ & OBZ$\uparrow$ & OBO$\uparrow$ \\ \midrule
MSE & 5.109 & 0.273 & 0.215 & 0.532\\
\rowcolor{LightRed} $\mathbf{+}$MAE &$\mathbf{4.455}$ & $\mathbf{0.213}$ & $\mathbf{0.245}$& $\mathbf{0.563}$ \\
\bottomrule
\end{tabular}
}
\end{minipage}%
\vspace{-1em}
\end{table}

\noindent
\textbf{Do exemplars help in training?} We study the impact of using exemplars for training by directly replacing the cross-attention decoder blocks with self-attention. 
As seen in~\cref{tab:self-attention_Vs_ours}, using self-attention (SA-only) performs significantly worse than our proposed ESCounts. Cross-attending exemplars decrease the RMSE/MAE by $-1.20$ and $-0.06$ whilst improving OBZ and OBO by $+0.10$ and $+0.09$, respectively. This emphasises the benefits of exemplar-based VRC. 

\noindent
\textbf{How many exemplars to sample?} A varying number of training exemplars $|\mathcal{S}|$ is used in~\cref{tab:multishot_in_training}. For $| \mathcal{S} | =0$, we train \textbf{only} the zero-shot latent $\mathbf{z}_0$ alongside the model's parameters. Training with $|\mathcal{S}| \sim \{0,1,2\}$ provides the best zero-shot scores at inference with our method efficiently learning to generalise by attending to only a few exemplars. The inclusion of more exemplars saturates performance.

\noindent
\textbf{How to sample exemplars?} 
In~\cref{tab:class_sampling} we analyse the impact of sampling exemplars from the same or other training videos. As expected, keeping the same action category for both exemplar and query videos performs the best, as ensuring the same action semantics between exemplars and query video helps to learn correspondence. In this table, we used sampling probability $p=0.4$.
In~\cref{tab:sampling exemplars}, we vary the sampling probability from other videos of the same underlying action $p$. 
For $p=0.0$, exemplars are sampled exclusively from the query video, whilst for $p=1.0$, exemplars are sampled solely from other videos of the same class. 
The best performance was observed with $p=0.4$, showcasing that the visual characteristics of exemplars from the same video are critical for VRC compared to class semantics.

\begin{figure}[tb]
\centering
\tabskip=0pt
\valign{#\cr
  \hbox{%
    \begin{subfigure}[b]{.21\textwidth}
    \centering
    \vspace{1em}
    \includegraphics[width=\linewidth]{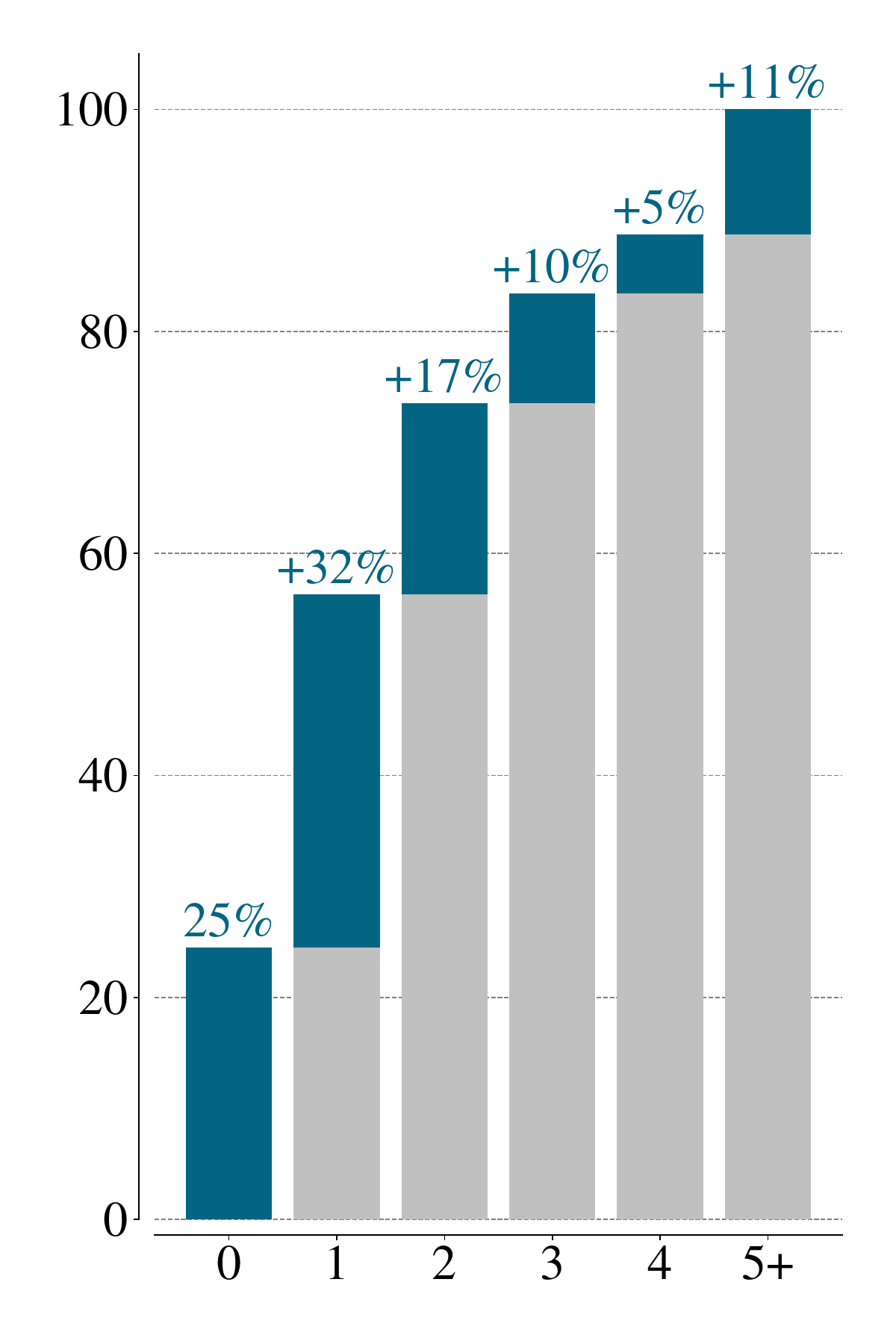}
    \caption{Off by N (OBN)}\label{fig:off-by-n}
    \end{subfigure}%
    \put (-73.0,55.0){\rotatebox{90}{\tiny{$\%$ of videos}}}
    }\vfill
    \hbox{%
    \begin{subfigure}[b]{.21\textwidth}
    \centering
    \includegraphics[width=\linewidth]{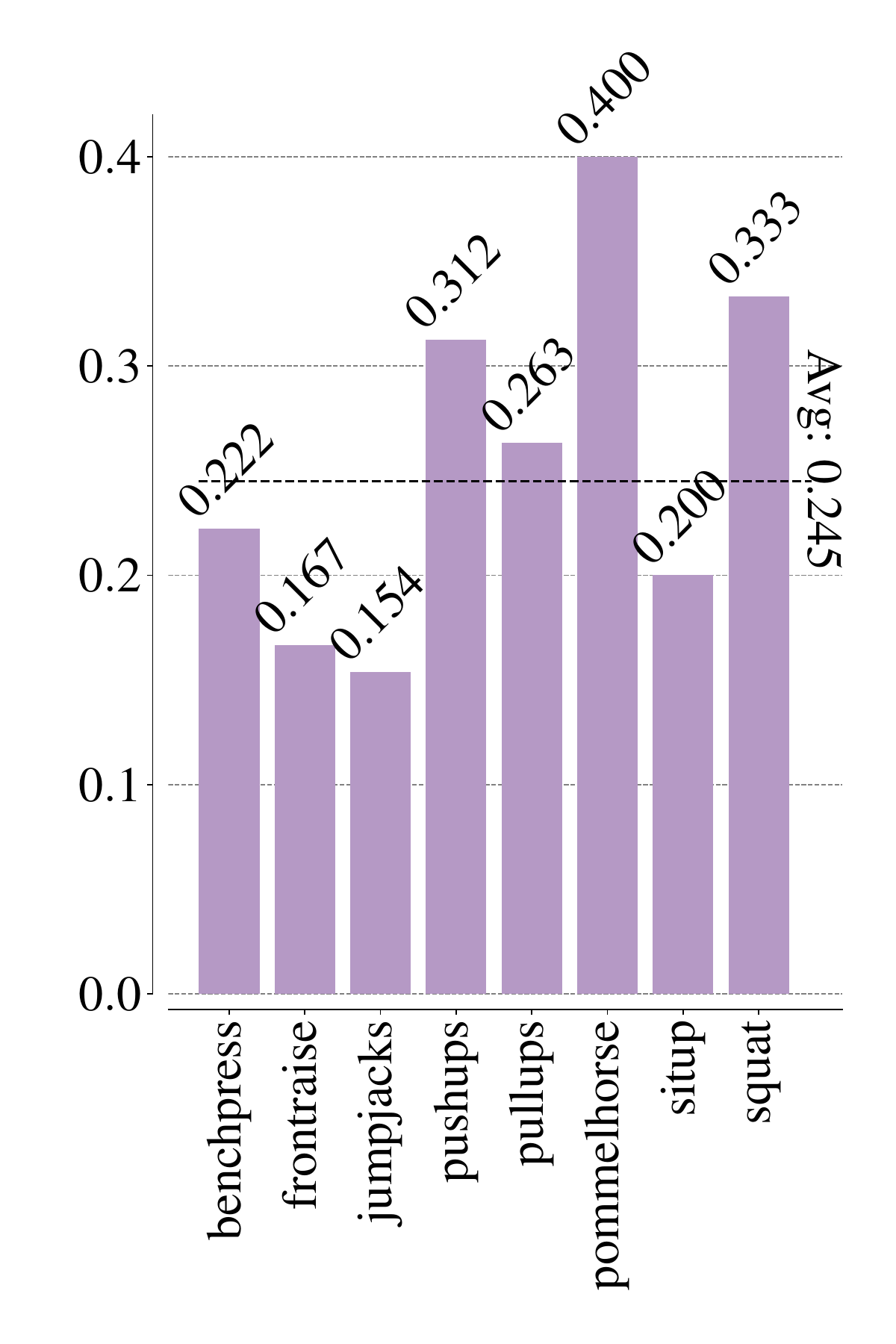}
    \caption{Class-wise OBZ}\label{fig:clswise_obz}
    \end{subfigure}%
    \put (-73.0,55.0){\rotatebox{90}{\tiny{OBZ}}}}\cr
  \noalign{\hfill}
  \hbox{%
    \begin{subfigure}{.195\textwidth}
    \centering
    \includegraphics[width=\linewidth]{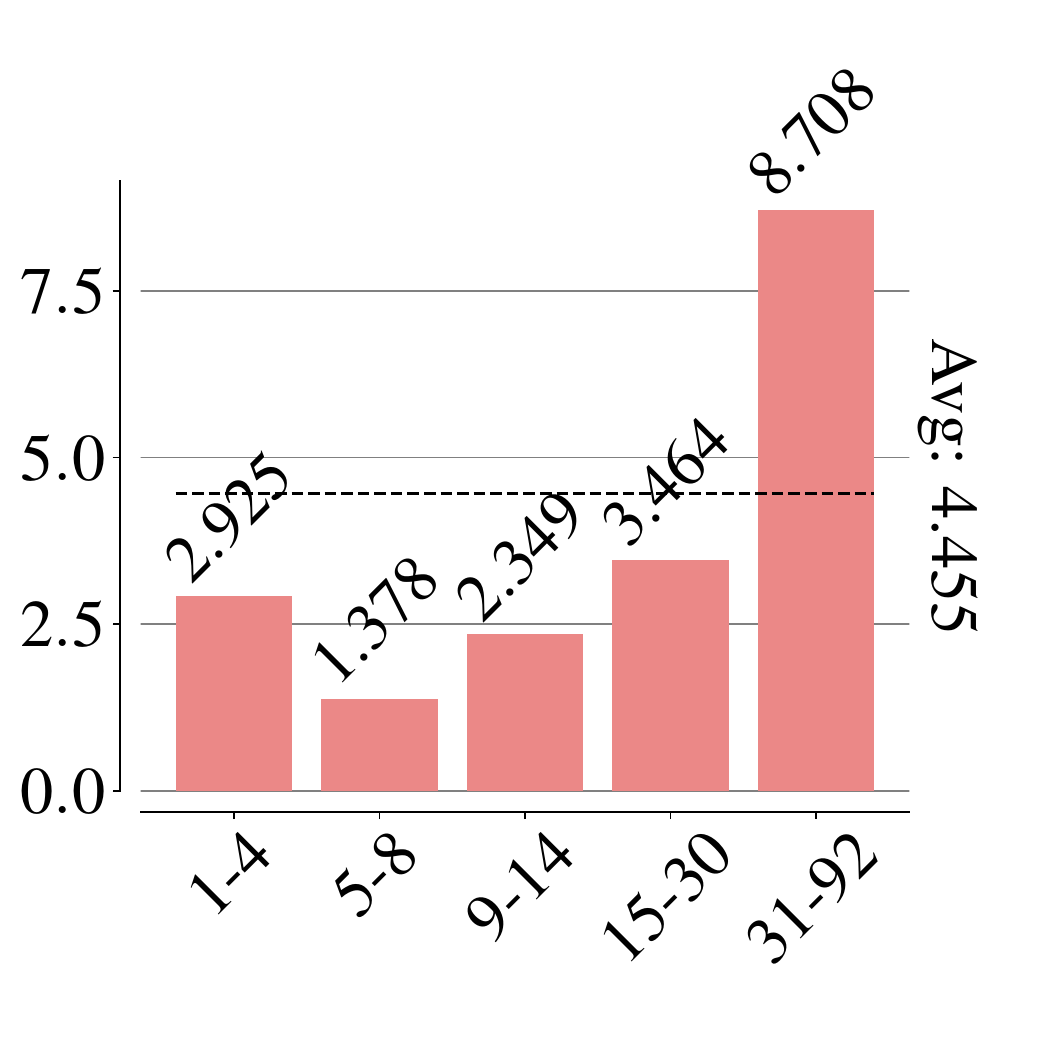}\vspace{.4em}
    \caption{RMSE}\label{fig:rmse-count}
    \end{subfigure}%
    \begin{subfigure}{.195\textwidth}
    \centering
    \includegraphics[width=\linewidth]{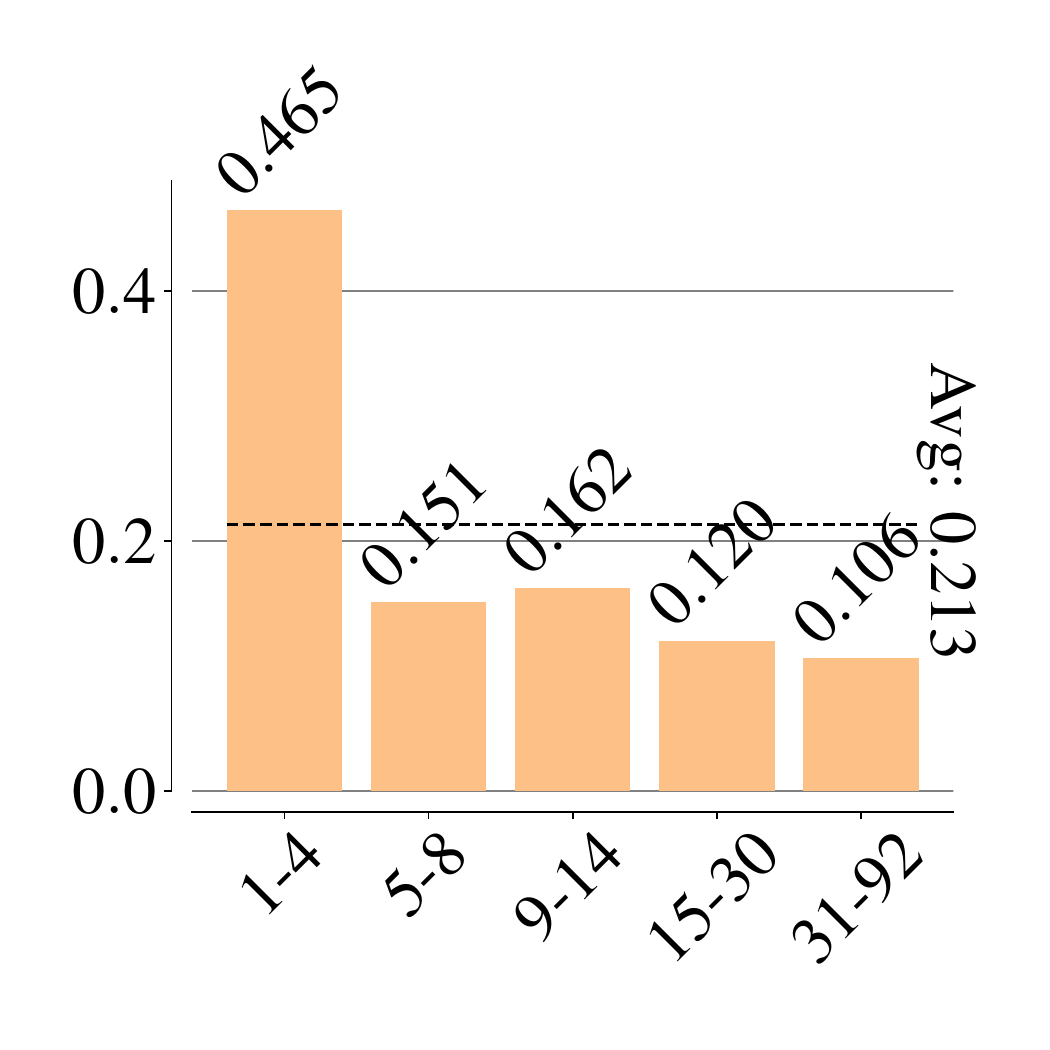}\vspace{.4em}
    \caption{MAE}\label{fig:mae-count}
    \end{subfigure}%
    \begin{subfigure}{.195\textwidth}
    \centering
    \includegraphics[width=\linewidth]{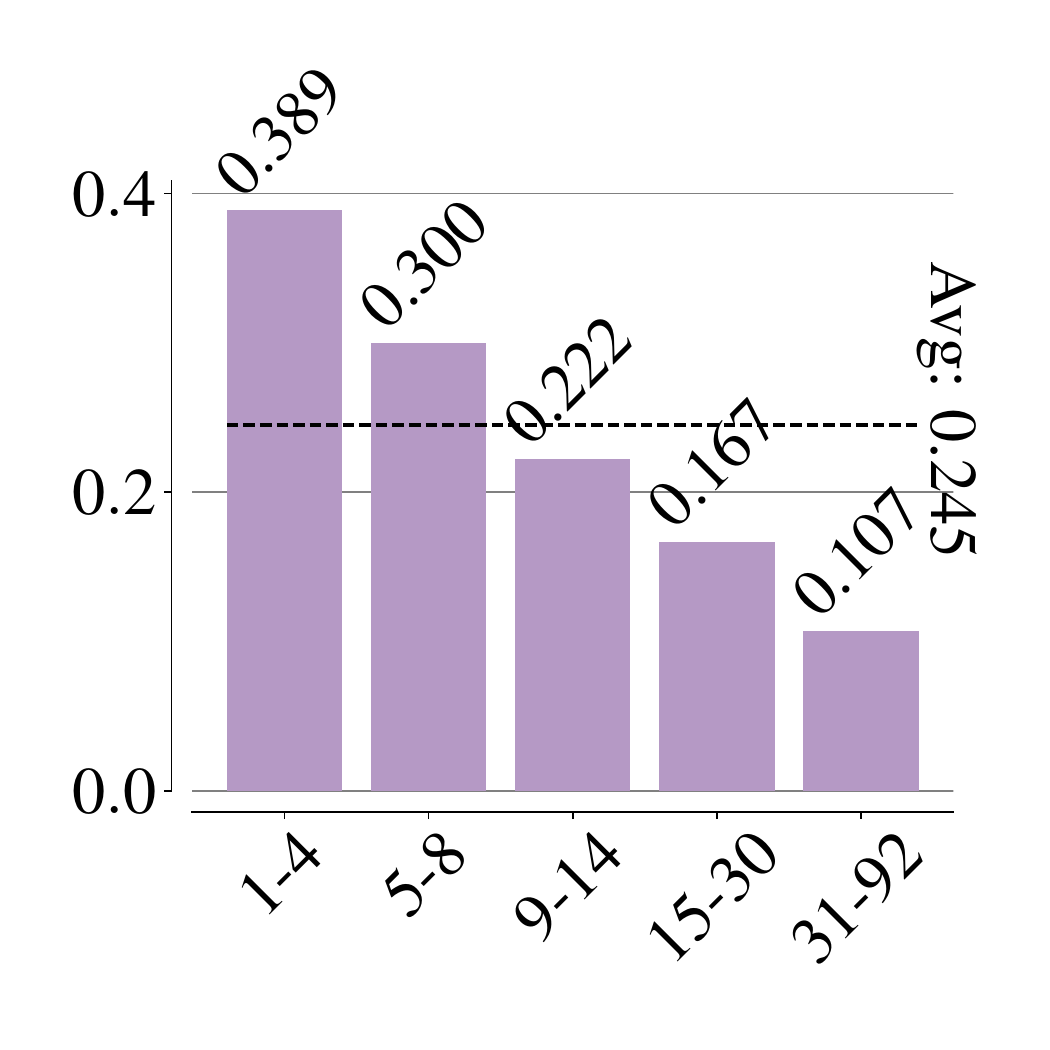}\vspace{.4em}
    \caption{OBZ}\label{fig:obz-count}
    \end{subfigure}%
    \begin{subfigure}{.195\textwidth}
    \centering
    \includegraphics[width=\linewidth]{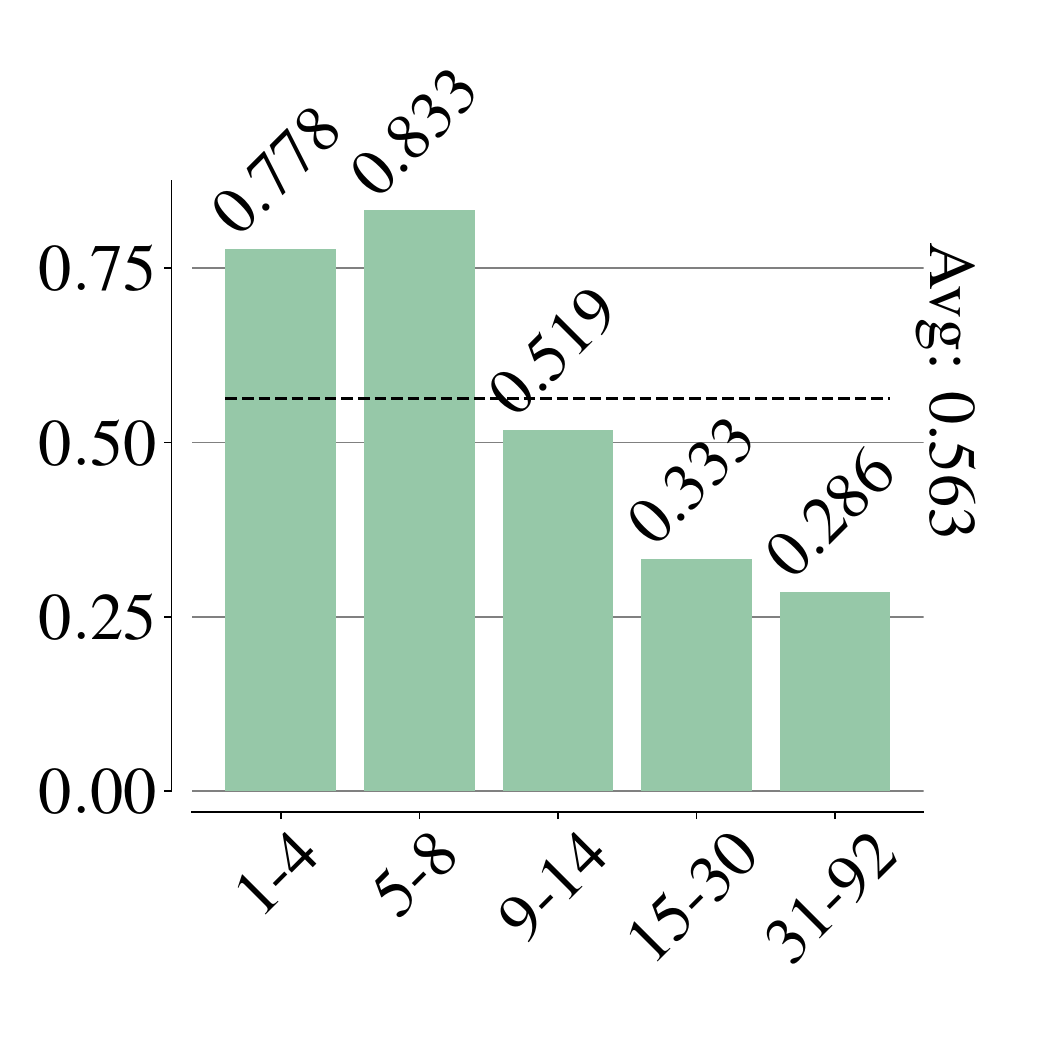}\vspace{.4em}
    \caption{OBO}\label{fig:obo-count}
    \end{subfigure}%
    \put (-190.0,13.0){\tiny{Ground truth number of repetitions}}
  
    }\vfill
    \hbox{%
    \begin{subfigure}{.195\textwidth}
    \centering
    \includegraphics[width=\linewidth]{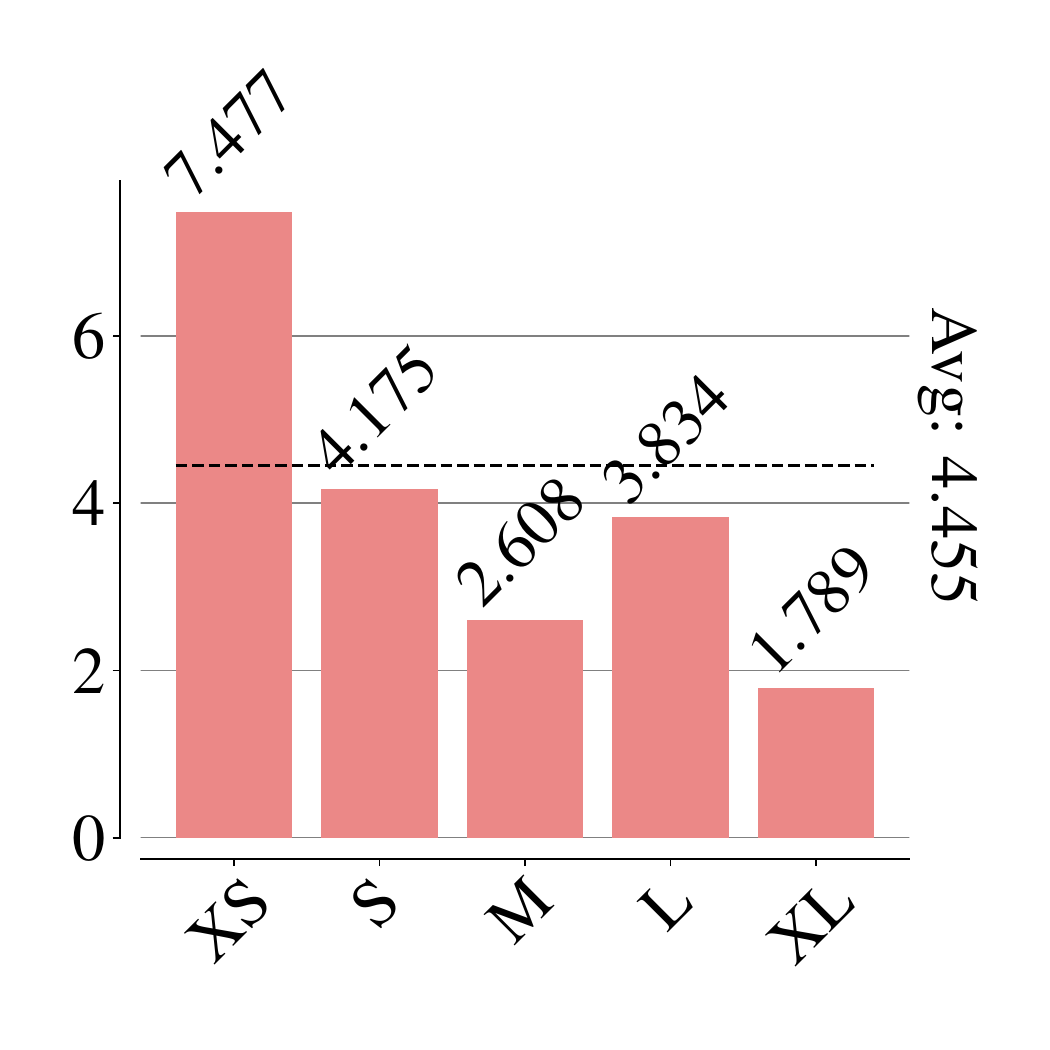}\vspace{.2em}
    \caption{RMSE}\label{fig:rmse-duration}
    \end{subfigure}%
    \begin{subfigure}{.195\textwidth}
    \centering
    \includegraphics[width=\linewidth]{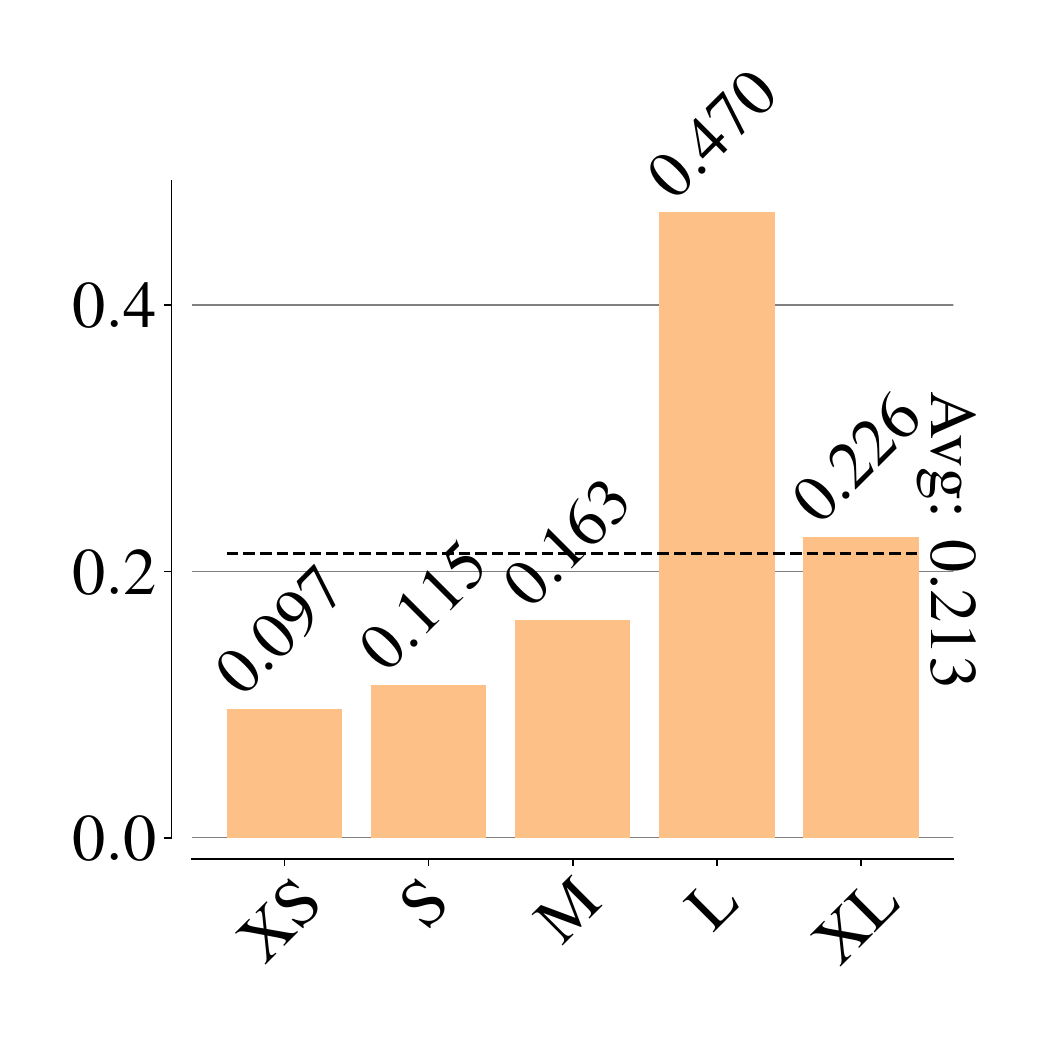}\vspace{.4em}
    \caption{MAE}\label{fig:mae-duration}
    \end{subfigure}%
    \begin{subfigure}{.195\textwidth}
    \centering
    \includegraphics[width=\linewidth]{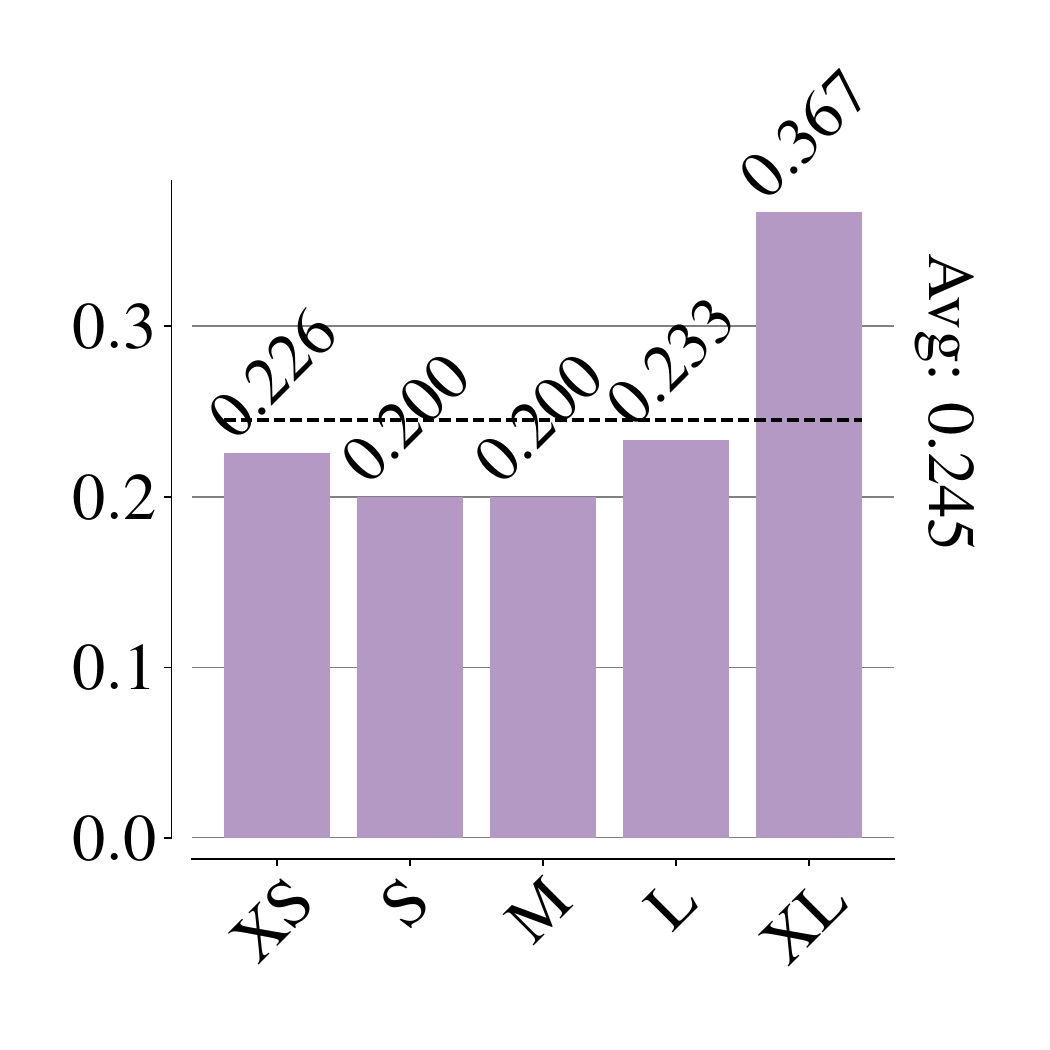}\vspace{.4em}
    \caption{OBZ}\label{fig:obz-duration}
    \end{subfigure}%
    \begin{subfigure}{.195\textwidth}
    \centering
    \includegraphics[width=\linewidth]{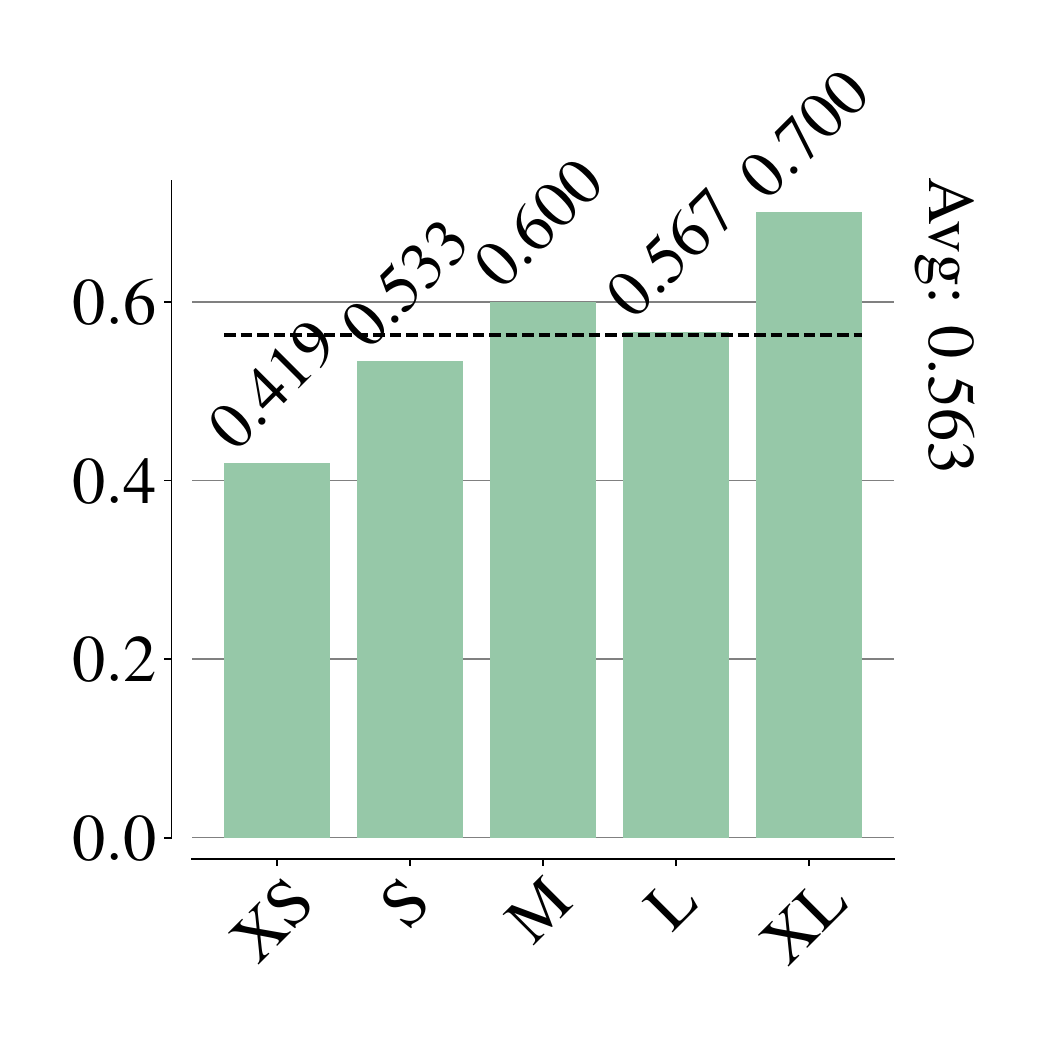}\vspace{.4em}
    \caption{OBO}\label{fig:obo-duration}
    \end{subfigure}%
    \put (-164.0,13.0){\tiny{Repetition length}}
    }\vfill
  \hbox{%
    \begin{subfigure}{.195\textwidth}
    \centering
    \includegraphics[width=\linewidth]{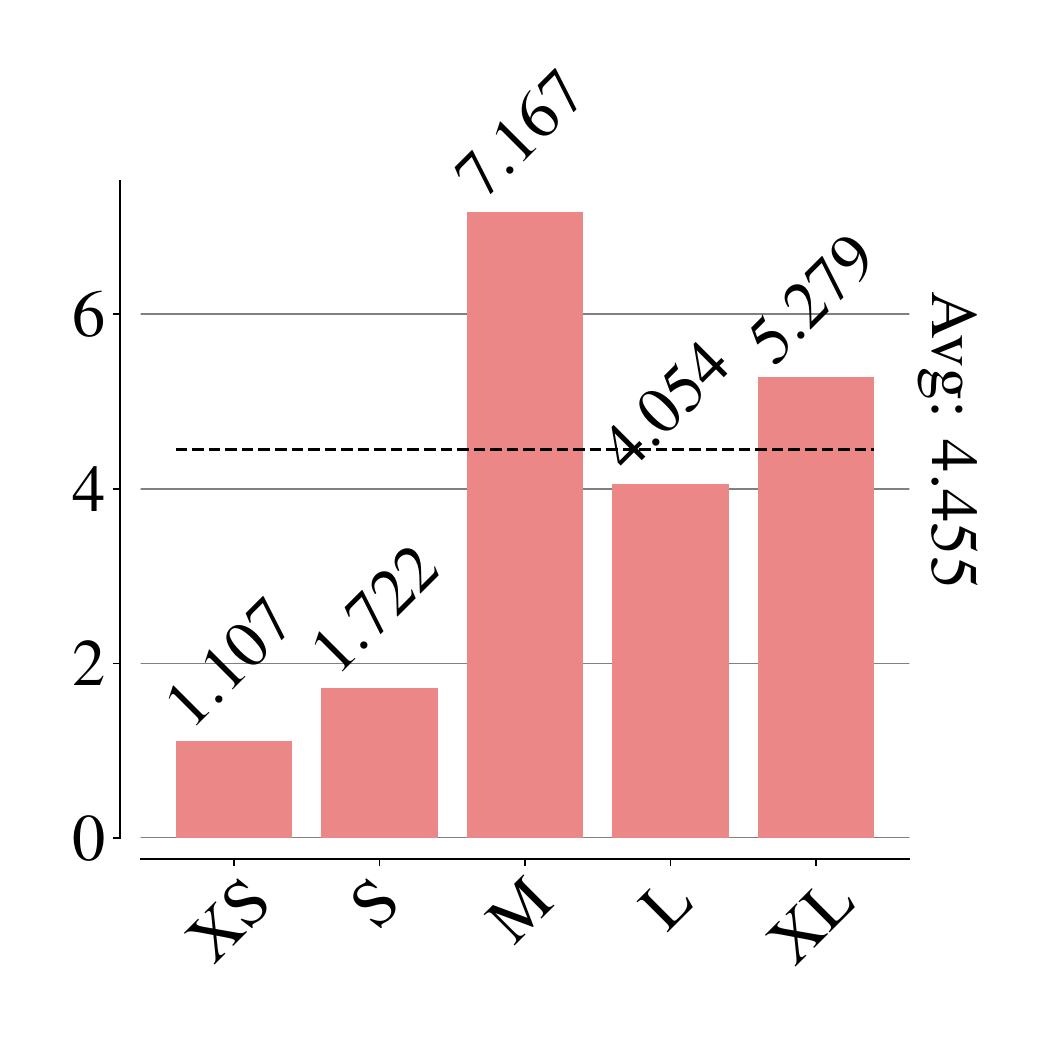}\vspace{.4em}
    \caption{RMSE}\label{fig:rmse-length}
    \end{subfigure}%
    \begin{subfigure}{.195\textwidth}
    \centering
    \includegraphics[width=\linewidth]{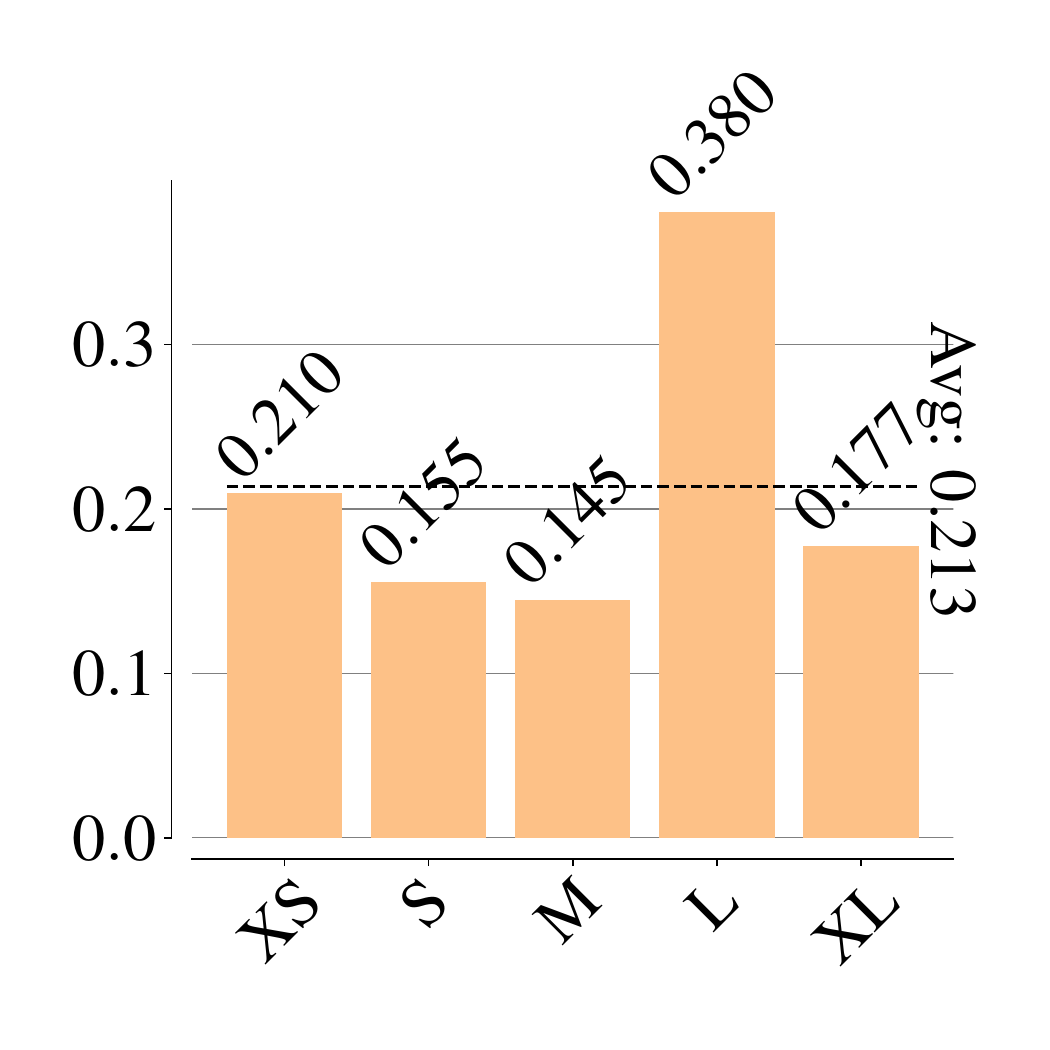}\vspace{.4em}
    \caption{MAE}\label{fig:mae-length}
    \end{subfigure}%
    \begin{subfigure}{.195\textwidth}
    \centering
    \includegraphics[width=\linewidth]{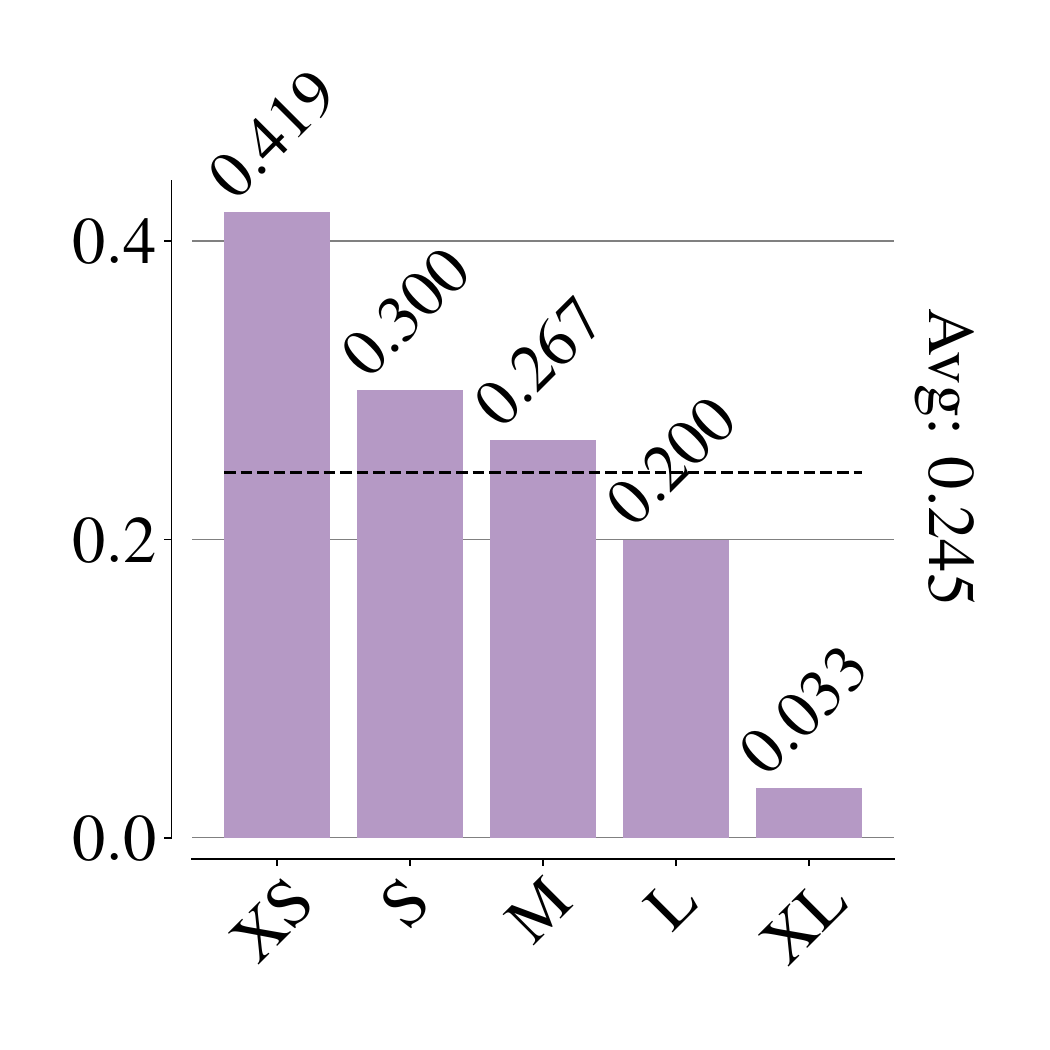}\vspace{.2em}
    \caption{OBZ}\label{fig:obz-length}
    \end{subfigure}%
    \begin{subfigure}{.195\textwidth}
    \centering
    \includegraphics[width=\linewidth]{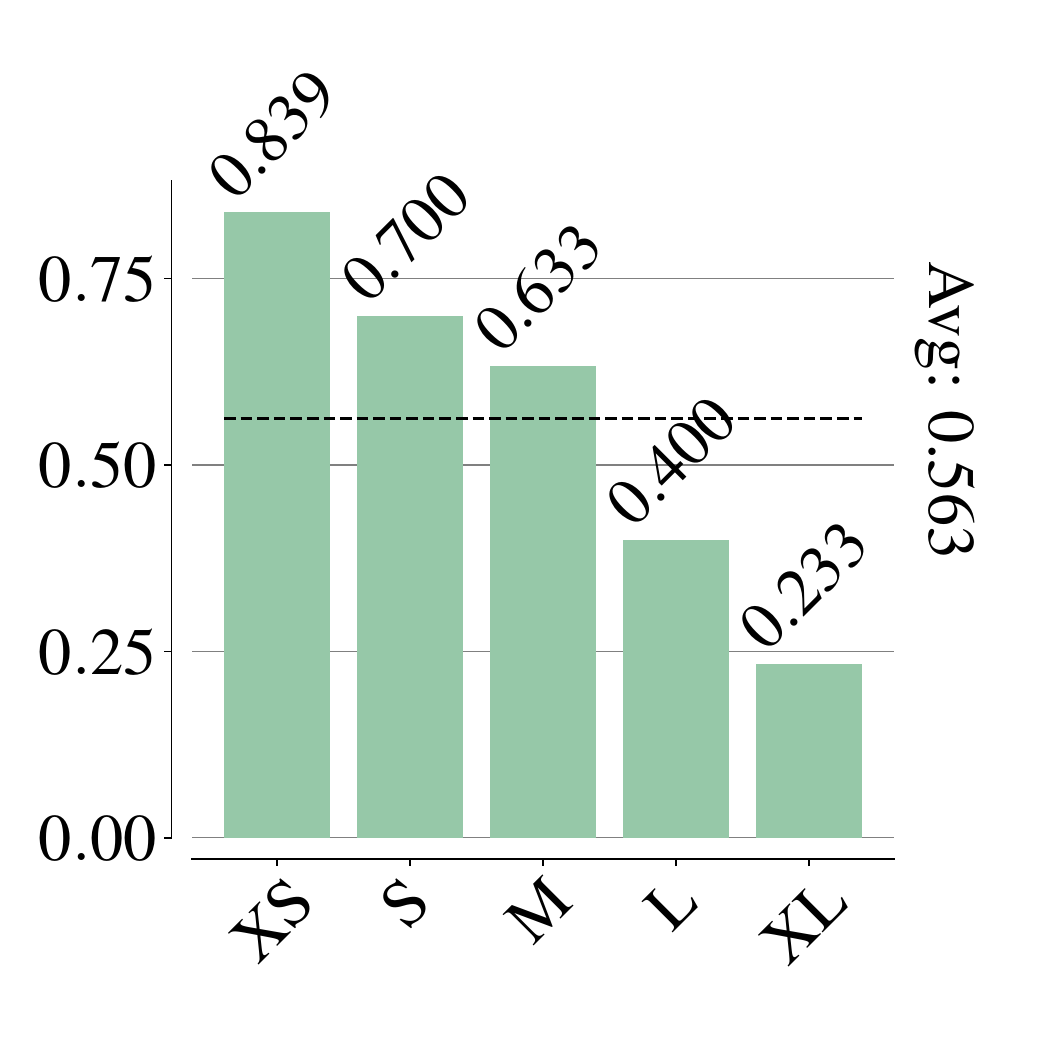}\vspace{.4em}
    \caption{OBO}\label{fig:obo-length}
    \end{subfigure}%
    \put (-164.0,13.0){\tiny{Video duration}}
  }\cr
}

\caption{\textbf{Grouped VRC scores} over different number of repetitions and lengths. 
(a)~overviews the Off by N 
accuracy for increasing Ns. (b) shows OBZ by action class. The first row (c--f) reports results
over different counts. (g--j) reports scores over groups by repetition durations. (k--n) reports metrics grouped by video duration. }
\label{fig:grouped_vrc_scores}
\vspace{-1.5em}
\end{figure}

\noindent
\textbf{What's the impact of time-shift augmentations?} 
Predictions are aggregated over $|\mathbf{K}|$ density maps by time-shifting the video input. As shown in~\cref{tab:use_overlapping_sequences}, having $|\mathbf{K}|=4$ shifted start/end positions provides the best results. However, results are strong even without test-time augmentations in $|\mathbf{K}|=1$.  

\noindent
\textbf{What should be the density map variance?} Density maps are constructed as vectors with normal distributions $\mathcal{N}(\cdot;\mu,\sigma)$ over repetition starts/ends timestamps. Reducing $\sigma$ increases the sharpness, resulting in a single delta function for $\sigma=0$. We ablate over different $\sigma$ in~\cref{tab:varying_width_of_density_peaks}. Denser and successive repetitions can benefit from sharper peaks of small $\sigma$ and sparser repetitions of larger durations can benefit from large $\sigma$. We also ablate using variable $\sigma$ that changes with the duration of repetition segments. Having $\sigma=0.5$ provides the best results with a balance between sharpness and covering the duration of repetitions.

\noindent
\textbf{How helpful is the MAE for the objective?}
We analyse ESCounts' performance with and without the MAE loss from~\cite{zhang2021repetitive} in \cref{tab:effect_of_Ls}. The combined objective helps performance for diverse counts across all metrics.

\noindent
\textbf{How close are predictions to the ground truth?} We further relax the off-by metrics to Off-By-N in~\cref{fig:off-by-n} to visualise the proximity of predictions to the ground truth. Overall, 84\% of predictions are within $\pm 3$ of the actual count.

\noindent
\textbf{What is the performance per action category?} In~\cref{fig:clswise_obz}, we plot the OBZ per action class. ESCounts performs fairly uniformly across all classes with the best-performing categories being \textit{pommelhourse} and \textit{squat}.

\noindent
\textbf{How does performance differ across counts, repetition lengths, and video durations?}
Up to this point, we have focused on the performance across all videos regardless of individual attributes. We now consider the sensitivity of ESCounts across equally sized groups based on the number of repetitions, average repetition length, and video duration. 

We report all metrics over groups of counts in~\cref{fig:rmse-count,fig:mae-count,fig:obz-count,fig:obo-count}. As expected, our method performs best in groups of smaller counts with higher counts being more challenging to predict precisely.

In~\cref{fig:rmse-duration,fig:mae-duration,fig:obz-duration,fig:obo-duration} we report VRC metrics with results grouped by average video repetition duration. These are grouped, into equal sized bins, to XS=(0-0.96)s, S=(0.96-1.53)s, M=(1.53-2.29)s, L=(2.29-3.09)s, XL=(>3.09)s. Predicting density maps is more challenging for short repetitions. However, ESCounts can still correctly predict counts across repetition lengths as shown by~\cref{fig:obz-duration,fig:obo-duration}. 
We also group videos by duration into XS=(8.0-11.0)s, S=(11.0, 26.0)s, M=(26.0, 33.9)s, L=(33.9-45.9)s and XL=(45.9-68.0)s. From~\cref{fig:rmse-length,fig:mae-length,fig:obz-length,fig:obo-length}, counting repetitions from longer videos is more challenging. 

\begin{wraptable}{r}{0.59\linewidth}
\caption{\textbf{Number of shots at inference}. We test using exemplars from the same video or a different video of the same action class from the train set.}
\centering
\resizebox{\linewidth}{!}{%
\begin{tabular}{cc cccc c cccc}
\toprule
& & \multicolumn{4}{c}{RepCount} & $\,$ & \multicolumn{4}{c}{UCFRep} \bstrut \\ \cline{3-6} \cline{8-11}
\multirow{-2}{*}{Shots} & \multirow{-2}{*}{\parbox{1cm}{Same video}} & RMSE$\downarrow$ & MAE$\downarrow$ & OBZ$\uparrow$ & OBO$\uparrow$ & & RMSE$\downarrow$ & MAE$\downarrow$ & OBZ$\uparrow$ & OBO$\uparrow$ \tstrut \bstrut \\ \midrule
\rowcolor{LightRed} 0 & N/A & 4.455 & 0.213 & 0.245 & 0.563 & & 1.972& 0.216& 0.381 & 0.704\\
\midrule
\multirow{2}{*}{1} & \ding{55}&4.432 & 0.207 & 0.251 & 0.563 & & 1.912 & 0.211 & 0.388 & 0.712\\
 & \ding{52}&4.369 & 0.210 & 0.247 & 0.589 & & 1.890 & 0.203 & 0.400 & 0.714 \\
\midrule
\multirow{2}{*}{2} &\ding{55} & 4.384& $\mathbf{0.206}$ & 0.251 & 0.572 & & 1.885 & 0.208 & 0.391 & 0.720 \\
 &\ding{52} & 4.360& 0.209 & 0.247 & 0.592 & & 1.857& 0.199 &  0.419 & 0.718\\
\midrule
\multirow{2}{*}{3} & \ding{55} &4.381 & 0.207 & $\mathbf{0.252}$& 0.579 & & 1.878 & 0.207 & 0.399 & $\mathbf{0.730}$\\ 
 & \ding{52} &$\mathbf{4.351}$ & $\mathbf{0.206}$ & 0.250 &  $\mathbf{0.596}$ & &$\mathbf{1.855}$ & $\mathbf{0.198}$&$\mathbf{0.420}$ & 0.723\\ 
\bottomrule
\end{tabular}%
}
\vspace{-4em}
\label{tab:multi-shot inference}
\end{wraptable}
\vspace{-2em}

\noindent
\subsection{Multi-Shot Inference } 
\label{sec:inference_with_more_shots} 
\vspace{-0.1em}
\looseness-1 We use learnt latents for exemplar-free inference.

Prior object counting~\cite{countr,lu2019class} report results with exemplars (i.e. object crops) at inference. 
While this is not comparable to other VRC works, we can assess our method's ability to utilise exemplars during inference in~\cref{tab:multi-shot inference}. Video exemplars steadily improve performance as the number of exemplars increases. Our model cross-attends exemplars in parallel, training with $0\!\!-\!\!2$ exemplars, and can even use $>\!2$ exemplars at inference. 
We show comparable results when sampling exemplars from the test video or training videos with the same action category. Combined with a classifier, a closed-set approach can be envisaged that classifies the action and then sources exemplars from the training set to assist counting during inference.

\section{Conclusion}

We have proposed to utilise exemplars for video repetition counting. We introduce Every Shot Counts (ESCounts), an attention-based encoder-decoder that learns to correspond exemplar repetitions across a full video. We define a learnable zero-shot latent that learns representations of generic repetitions, to use during inference. Extensive evaluation on RepCount, Countix, and UCFRep demonstrates the merits of ESCounts achieving state-of-the-art results on the traditional MAE and OBO metrics and the newly introduced RMSE and OBZ. 
We provide detailed analysis and ablations of our method, highlighting the importance of training with exemplars and time-shift augmentations.
The diversity of these exemplars is an aspect for future exploration.

\noindent \textbf{Acknowledgements.}
This work uses publicly available datasets and annotations for results and ablations.
Research is supported by EPSRC UMPIRE (EP/T004991/1). S.~Sinha is supported by EPSRC DTP studentship.

%
%
\bibliographystyle{splncs04}
\bibliography{main}

\setcounter{section}{5}
\setcounter{equation}{11}
\setcounter{figure}{6}
\setcounter{table}{4}

\newpage
\section*{Appendix}

Code is made publicly available at: \url{https://github.com/sinhasaptarshi/EveryShotCounts}. The repository contains the full train and evaluation code and a demo for inference with a few videos.

In the following 
sections, we 
provide more qualitative results 
in~\cref{sec:video}. We then provide additional ablations on the architecture's choices (e.g. depth of transformer and window size) in~\cref{sec:ablations2}.
Additionally, we evaluate the ability of ESCounts to locate each repetition within the video in~\cref{sec:localisation_results}.
We then compare VRC to Temporal Action Segmentation (TAS) in~\cref{sec:vrc2tas} demonstrating distinctions between the two tasks. 

Additionally, following the release of the recent egocentric video counting dataset OVR-Ego4D \cite{dwibedi2024ovr}, we train and evaluate ESCounts on this newly introduced dataset demonstrating the effectiveness of our method for egocentric counting in \cref{sec:egoresults}.

\begin{table}[h]
\begin{minipage}[t]{0.30\textwidth}

    \centering
    \caption{\textbf{Impact of} $L$.}
    \resizebox{\linewidth}{!}{
    \begin{tabular}{ccccc}
    \toprule
        $L$ & RMSE$\downarrow$ & MAE$\downarrow$ & OBZ$\uparrow$ & OBO$\uparrow$ \\
        \midrule
         1&4.843&0.229&0.223&0.545 \\
         \rowcolor{LightRed}2&$\mathbf{4.455}$&$\mathbf{0.213}$&0.245& $\mathbf{0.563}$\\
         3 &4.575&0.219&$\mathbf{0.247}$&0.560 \\
         4 &4.783&0.225&0.235& 0.548\\
         \bottomrule
    \end{tabular}
    }
    \label{tab:impact_of_L}
\end{minipage}%
\hfill
\begin{minipage}[t]{0.30\textwidth}
    \centering
    \caption{\textbf{Impact of} $L'$.}
    \resizebox{\linewidth}{!}{
    \begin{tabular}{ccccc}
    \toprule
        $L'$ & RMSE$\downarrow$ & MAE$\downarrow$ & OBZ$\uparrow$ & OBO$\uparrow$ \\
        \midrule
         1&4.932&0.247&0.212& 0.525\\
         2&4.634&0.218&0.238&0.550 \\
         \rowcolor{LightRed} 3 &$\mathbf{4.455}$&$\mathbf{0.213}$&$\mathbf{0.245}$& $\mathbf{0.563}$\\
         4 &4.532&0.225&0.230& 0.552\\
         \bottomrule
    \end{tabular}}
    \label{tab:impact_of_L'}
    \end{minipage}%
    \hfill
\begin{minipage}[t]{0.38\textwidth}
    \centering
    \caption{\textbf{Window sizes}.}
    \resizebox{0.9\linewidth}{!}{
    \begin{tabular}{ccccc}
    \toprule
        $(t',h',w')$ & RMSE$\downarrow$ & MAE$\downarrow$ & OBZ$\uparrow$ & OBO$\uparrow$ \\
        \midrule
         $(3,3,3)$&5.212&0.261&0.185& 0.521\\
        $(2,7,7)$&4.871&0.247&0.201& 0.537\\
        \rowcolor{LightRed}$(4,7,7)$&$\mathbf{4.455}$&$\mathbf{0.213}$&$\mathbf{0.245}$& $\mathbf{0.563}$\\
        $(7,7,7)$&4.753&0.225&0.232& 0.520\\
        \textit{full} & 5.011 & 0.227 & 0.221 & 0.533\\
         \bottomrule
    \end{tabular}}
    \label{tab:self_attention_vs_windowedself_attention}
    \end{minipage}
\end{table}

\section{Qualitative Video and Extended Figure}
\label{sec:video}
We provide a compilation of videos on our website \url{https://sinhasaptarshi.github.io/escounts/} showcasing our method's Video Repetition Counting (VRC) abilities over a diverse set of 20 videos from all 3 datasets. Videos are shown alongside synchronised ground truth and predicted density maps. The test set from which each video is sampled is also shown.

We additionally extend~\cref{fig:qualitative} in the main paper with more examples from all datasets in~\cref{fig:more_qualitative}.

\section{Further Ablations}
\label{sec:ablations2}
\looseness-1 We extend the ablations in~\cref{sec:ablation_studies}, report results over different $L$ and $L'$, and analyse the impact of windowed-self attention on the performance of ESCounts.

\begin{figure}[tp!]
\begin{minipage}[c]{\textwidth}
    \centering
    \includegraphics[width=\textwidth,trim={2.5cm 0 3cm 0},clip]{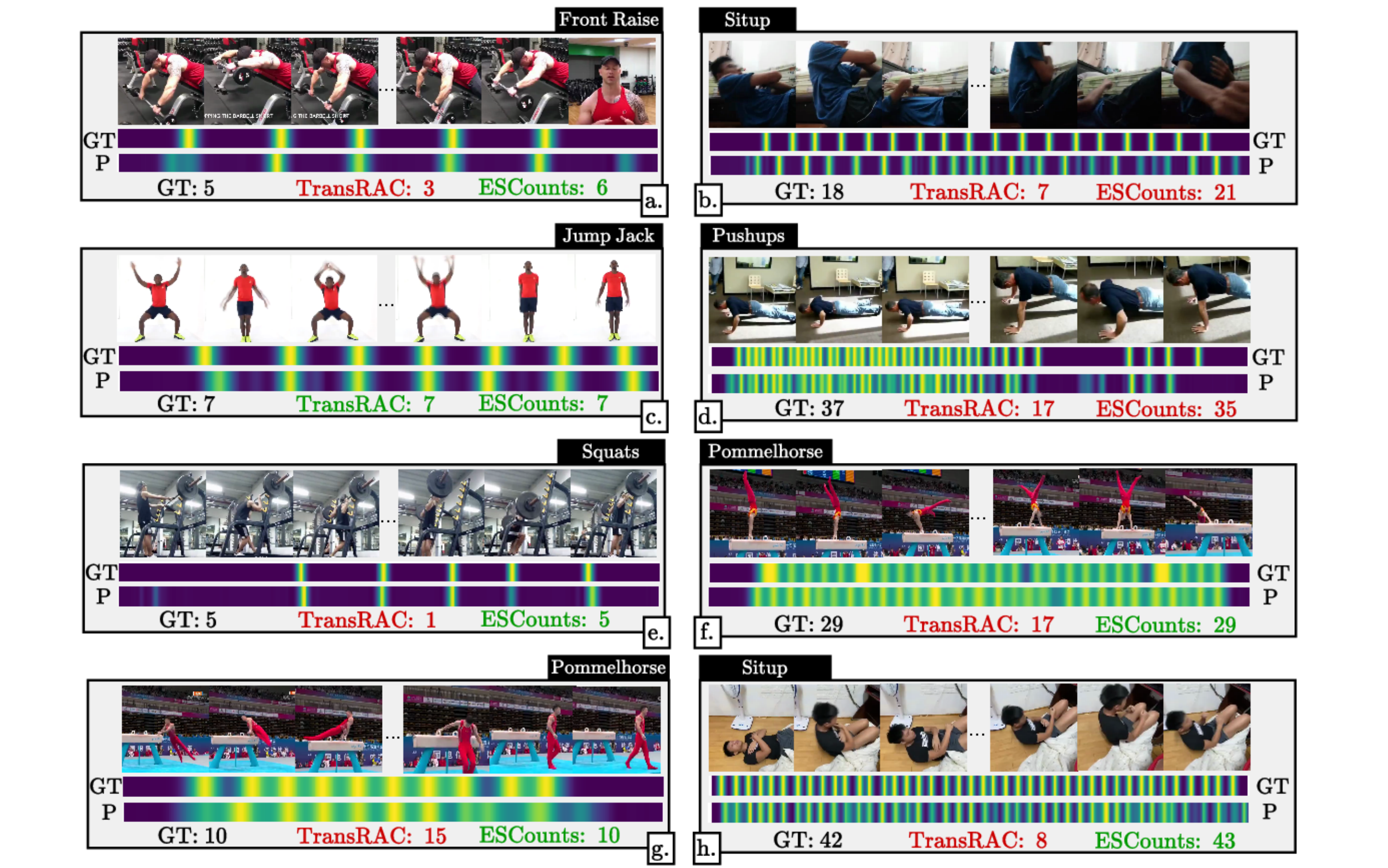}
    \subcaption{RepCount}
    \label{fig:more_qualitative_repcount}
\end{minipage}
\begin{minipage}[c]{\textwidth}
\vspace{2em}
    \centering
\includegraphics[width=\textwidth,trim={2.5cm 0 2.5cm 0},clip]{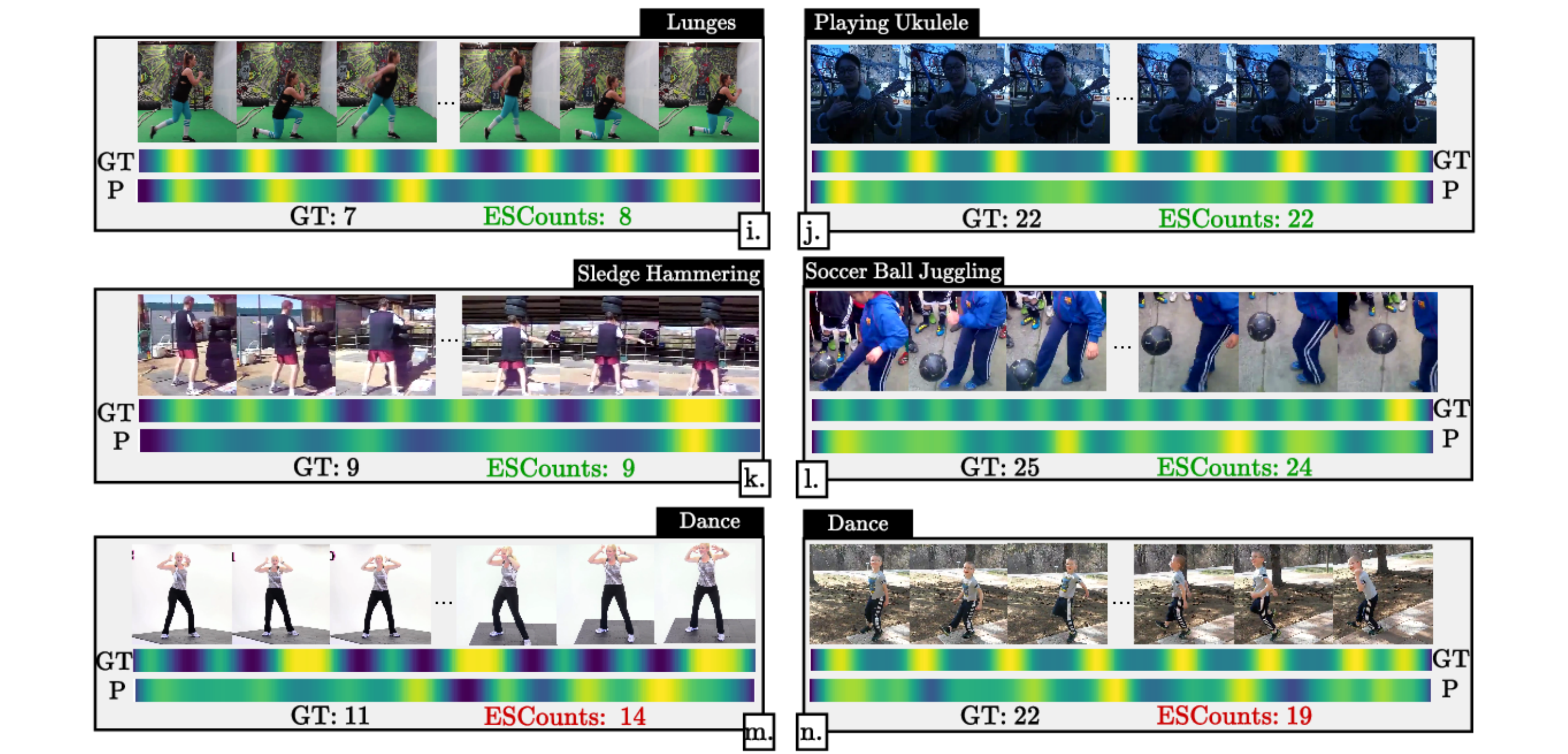}
    \subcaption{ Countix}
\label{fig:more_qualitative_countix}
\end{minipage}
\caption{\textbf{Additional qualitative results}.}
\end{figure}%
\begin{figure}[tp!]\ContinuedFloat
\begin{minipage}[c]{\textwidth}
    \centering
    \includegraphics[width=\textwidth,trim={2.5cm 0 2.5cm 0},clip]{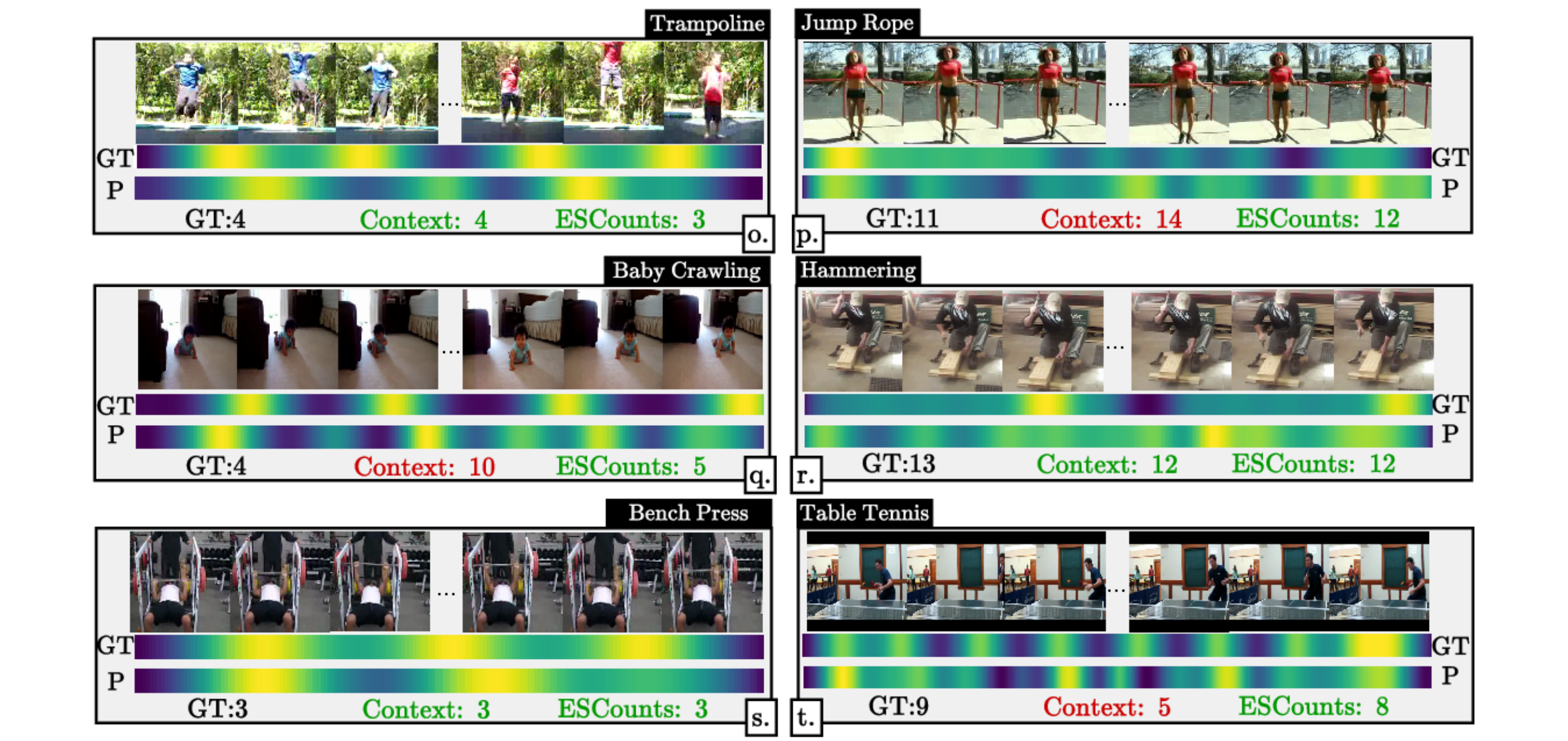}
    \subcaption{ UCFRep}
\label{fig:more_qualitative_ucfrep}
\end{minipage}
\caption{\textbf{Additional qualitative results} (continued).}
\label{fig:more_qualitative}
\end{figure}

\noindent
\textbf{Impact of $L$. }
We ablate $L$ \ie the number of layers in the cross-attention block. Increasing $L$ increases the number of operations that discover correspondences between the video and the selected exemplars. As seen in~\cref{tab:impact_of_L}, while low $L$ causes a drop in performance, high $L$ can also be detrimental probably due to overfitting. $L=2$ gives the best results for the majority of the metrics. 

Next keeping $L=2$ fixed, we vary $L'$ in~\cref{tab:impact_of_L'}. $L'$ is the number of windowed self-attention layers in the self-attention block. $L'=3$ gives the best results across all the metrics. Similarly, increasing or decreasing $L'$ drops performance gradually.

\noindent
\textbf{Self-attention vs Windowed Self-attention}. Motivated by~\cite{liu2022video}, we use windowed self-attention for the decoder self-attention blocks. Given spatio-temporal tokens $\mathcal{T}' \times H' \times W' \times C$, windowed self-attention computes multi-headed attention for each token within the immediate neighbourhood using 3D shifted windows of size $t' \times h' \times w'$, where $t'\leq \mathcal{T}'$, $h'\leq H'$ and $w'\leq W'$. We ablate on various $(t', h', w')$ values in~\cref{tab:self_attention_vs_windowedself_attention}. Note that for $t'= \mathcal{T}'$, $h'= H'$, and $w'= W'$ denoted as \textit{full}, standard self-attention is used where each token attends to every token. As shown, the best performance is obtained with window size $(4,7,7)$, demonstrating the importance of attending to tokens in immediate spatio-temporal neighbourhoods only. 
We found variations in the value of $t'$ to have the largest performance impact with 
 decreases as the value of $t'$ changes.

\noindent
\textbf{Sampling Rate for Encoding.} As stated in the implementation details, we sample every four frames from the video to form the encoder inputs. We ablate the impact of the sampling rate in~\cref{tab:impact_of_samplingrate}. As shown, denser sampling is key for robust video repetition counting. Reducing the sampling rate steadily decreases performance as relevant parts of repetitions may be missed.   

\begin{table}[tp]
    \centering
    \caption{\textbf{Impact of sampling rate}}
    \begin{tabular}{ccccc}
    \toprule
         \begin{tabular}{c}Sampling every \\ $n$ frames\end{tabular} & RMSE $\downarrow$& MAE $\downarrow$ & OBZ$\uparrow$ & OBO$\uparrow$ \\
         \midrule
         \rowcolor{LightRed} 4 & \textbf{4.455} & \textbf{0.213} & \textbf{0.245} & \textbf{0.563} \\
         8 & 5.112 & 0.268& 0.221 & 0.521 \\
         16 & 5.911& 0.296 & 0.185 &0.482 \\
         32 & 6.562 & 0.346& 0.156 & 0.444 \\
         \bottomrule
    \end{tabular}
    \label{tab:impact_of_samplingrate}
\end{table}
\begin{table}
    \centering
    \caption{\textbf{OBO, parameters, and training and inference speeds on UCFRep}. Metrics obtained by the public available codebase of~\cite{zhang2020context} are denoted with $^{\textcolor{red}{*}}$.}
    \begin{tabular}{lc cc ll ll ll }
    \toprule
         Method & $\;$ &OBO$\uparrow$ & $\;$ & \begin{tabular}{c}\#Trainable \\ params (M)\end{tabular} & $\;$ & \begin{tabular}{c}Train set $\downarrow$\\ (sec/sample) \end{tabular} & $\;$  & \begin{tabular}{c}
              Test set $\downarrow$  \\
             (sec/sample)
         \end{tabular}\\
         \midrule
         Context\cite{zhang2020context}  && \textbf{0.790} && 47.6$^{\textcolor{red}{*}}$ && 1.171$^{\textcolor{red}{*}}$ && 1.818$^{\textcolor{red}{*}}$ \bstrut\\
        \rowcolor{LightRed} ESCounts &&0.731 && \textbf{21.1} \textcolor{applegreen}{(-26.5)} && \textbf{0.138} \textcolor{applegreen}{(-1.033)} && \textbf{0.141} \textcolor{applegreen}{(-1.677)}\\
         \bottomrule
\end{tabular}
\label{tab:speed}

\end{table}

\noindent
\textbf{Model Size and Speed}
For UCFRep \cite{zhang2020context}, \cite{zhang2020context,zhang2021repetitive} achieve better performance than ESCounts. However, this performance is achieved by having more trainable parameters, as~\cite{zhang2020context,zhang2021repetitive} finetune the encoders on the target dataset. We use the provided codebase from~\cite{zhang2020context} and benchmark the average number of iterations per second for a full forward and backward pass over the entire training set. Additionally, we report inference-only average times on the test set. We use the same experiment set-up described in Sec.~\textcolor{red}{4.1} and report speeds in~\cref{tab:speed}. 
Training ESCounts is $\sim$8$\times$ faster. Interestingly, ESCounts maintains its efficiency even during inference with $\sim$12$\times$ faster times than Context~\cite{zhang2020context} which uses iterative processing. Note that~\cite{zhang2021repetitive} could not be used for this analysis as their code for training with UCFRep is not publicly available.

\begin{table}[tp]
\caption{\textbf{Repetition  localisation results on RepCount} measured as the mAP ($\%$) over different Jaccard index relative thresholds $r$.}
\label{table:localise}
\centering
\resizebox{0.95\textwidth}{!}{%
\begin{tabular}{l c c c c c c c c c c c >{\columncolor[gray]{0.9}}c}
\hline
\multirow{2}{*}{Method} && \multicolumn{9}{c}{$\theta$ values for relative threshold $r$} && \tstrut \\
&& 0.1 & 0.2 & 0.3 & 0.4 & 0.5 & 0.6 & 0.7 & 0.8 & 0.9 && \multirow{-2}{*}{Avg} \bstrut \\
\hline
Baseline~\cite{hu2022tranrac} && 38.59 & 37.46 & 35.02 & 32.55 & 30.40 & 26.97 & 22.66 & 17.22 & 12.17 && 28.12 \tstrut \\
\rowcolor{LightRed} ESCounts && \textbf{38.83} & \textbf{38.64} & \textbf{38.07} & \textbf{37.44} & \textbf{35.82} & \textbf{33.43} & \textbf{30.76} & \textbf{27.52} & \textbf{20.85} && \textbf{33.48} \bstrut \\
\hline
\end{tabular}%
}
\end{table}

\begin{wrapfigure}{r}{0.42\textwidth}
    \vspace{-2em}
    \centering
    \includegraphics[width=\linewidth,trim={2cm 1cm 0cm 0},clip]{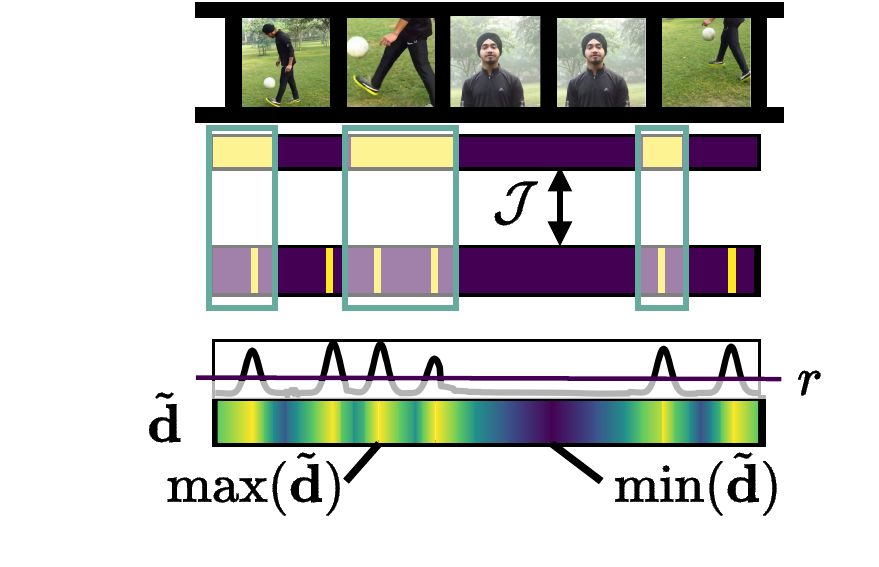}
    \caption{\textbf{Localisation metric $\mathcal{J}$}. 
    We identify local maxima in $\tilde{\mathbf{d}}$ and threshold peaks higher than $r$ to remove noise. $\mathcal{J}$ is then computed between the annotated start-end times and the thresholded peaks.}
    \label{fig:loc}
    \vspace{-2em}
\end{wrapfigure}
\section{Repetition Localisation}
\label{sec:localisation_results}

VRC metrics only relate predicted to correct counts, regardless of whether the repetitions have been correctly identified.
We thus investigate whether the peaks of the predicted density map $\tilde{\mathbf{d}}$ align with the annotated start-end times of repetitions
in the ground truth. Following action localisation 
methods~\cite{caba2015activitynet,gu2018ava,idrees2017thumos}, we adopt the Jaccard index $\mathcal{J}$ for repetition localisation. 
As the values of $\tilde{\mathbf{d}}$ peaks vary across videos, we apply thresholds $\theta$ relative to the maximum and minimum values, $r = \theta (\max(\tilde{\mathbf{d}}) - \min(\tilde{\mathbf{d}}))$. 
We find all local maxima in $\tilde{\mathbf{d}}$ and only keep those above threshold $r$.
We consider a repetition to be correctly located (TP) if at least one peak occurs within the start-end time of that repetition. Peaks that occur within the same repetition are counted as one. In contrast, peaks that do not overlap with repetitions are false positives (FP) and repetitions that do not overlap with any peak are false negatives (FN).
We then calculate $\mathcal{J}$ as TP divided by all the correspondences (TP + FP + FN) as customary.

In~\cref{table:localise} we report the Jaccard index over different thresholds alongside the Mean Average Precision (mAP) on RepCount. We select TransRAC~\cite{hu2022tranrac} as a baseline due to their publicly available checkpoint. Across thresholds, ESCounts outperforms~\cite{hu2022tranrac} with the most notable improvements observed over higher threshold values. This demonstrates ESCounts' ability to predict density maps with higher contrast between higher and lower salient regions. For 0.9, 0.8, and 0.7 thresholds ESCounts demonstrates a +8.68\%, +10.30\%, and +8.10\% improvement over~\cite{hu2022tranrac}.

\begin{table}
\centering
\caption{\textbf{Comparison between ESCounts and TAS baseline on close and open-set RepCount setting}.}
\begin{tabular}{lllllll}
\toprule
\multirow{2}{*}{Task} & \multirow{2}{*}{Method} & \multicolumn{2}{c}{benchmark} && \multicolumn{2}{c}{open-set}\bstrut \\ \cline{3-4}\cline{6-7}
 && MAE$\downarrow$ & OBO$\uparrow$ && MAE$\downarrow$ & OBO$\uparrow$ \tstrut \\
\midrule
\multirow{2}{*}{TAS} & GTRM~\cite{huang2020improving} & 0.527 & 0.159 &$\;$& 1.000 & 0.000 \\
& TriDet \cite{shi2023tridet} & 0.603 & 0.232 && 1.000 &0.000\\
\hline
\cellcolor{LightRed} VRC & \cellcolor{LightRed} ESCounts & \cellcolor{LightRed}\textbf{0.213}  & \cellcolor{LightRed}\textbf{0.563} & \cellcolor{LightRed} & \cellcolor{LightRed}\textbf{0.436} & \cellcolor{LightRed}\textbf{0.519} \tstrut \\
\bottomrule
\end{tabular}

\label{tab:vrc2tas}
\vspace{-1em}
\end{table}

\section{Distinction between VRC and TAS}
\label{sec:vrc2tas}

Unlike Temporal Action Segmentation (TAS) methods, VRC methods can generalise to unseen action classes. In~\cref{tab:vrc2tas} we compare ESCounts to a TAS method~\cite{huang2020improving} on the RepCount benchmark (\emph{close-set}) and \emph{open-set} setting. As shown, \cite{huang2020improving} can only localise the actions of a pre-defined set of categories with which the model was trained. In contrast, VRC is learned as an \emph{open-set} task. As ESCounts uses a learnt latent to encode class-independent repetition embeddings, it effectively generalises to unseen categories.  
In addition, ESCounts can better handle large variations in repetition durations that are present in VRC videos compared to~\cite{huang2020improving}, which as noted by~\cite{hu2022tranrac} is a weakness of TAS methods.

\section{Results on egocentric VRC.}
\label{sec:egoresults}

\begin{table}[h]
    \centering
    \caption{\textbf{Results on OVR-Ego4D}.$\dagger$ indicates results have been copied from \cite{dwibedi2024ovr}. (V) corresponds to vision-only models and (V+L) to vision and language models.}
    \label{tab:ovrego4d}
    \begin{tabular}{clcccc}
    \toprule
         Modality & Method & RMSE $\downarrow$& MAE $\downarrow$ & OBZ$\uparrow$ & OBO$\uparrow$ \\
         \midrule
         \multirow{2}{*}{V} & RepNet \cite{dwibedi2020counting} $\dagger$ & 3.20 & 0.74 & 0.19 & 0.43 \\
          & \cellcolor{LightRed} ESCounts & \cellcolor{LightRed} 2.41 & \cellcolor{LightRed} \textbf{0.32} & \cellcolor{LightRed} \textbf{0.30} & \cellcolor{LightRed} \textbf{0.68} \\
          \midrule
         V+L & OVRCounter \cite{dwibedi2024ovr} $\dagger$ & \textbf{1.60} & 0.35 & 0.29 & 0.66 \\
         \bottomrule
    \end{tabular}
\end{table}

The recently-introduced OVR-Ego4D \cite{dwibedi2024ovr} is an Ego4D \cite{Ego4D} subset containing clips of repetitive egocentric actions, \eg cutting onions, rolling dough. It comprises 50.6K 10-second clips with 41.9K train and 8.7K test clips. Annotations are only provided for the number of repetitions and not the individual start and end times per repetition. Thus, similar to Countix, we define pseudo-labels to estimate the density maps. 

\begin{figure}[tp!]
\begin{minipage}[c]{\textwidth}
    \centering
    \includegraphics[width=\textwidth]{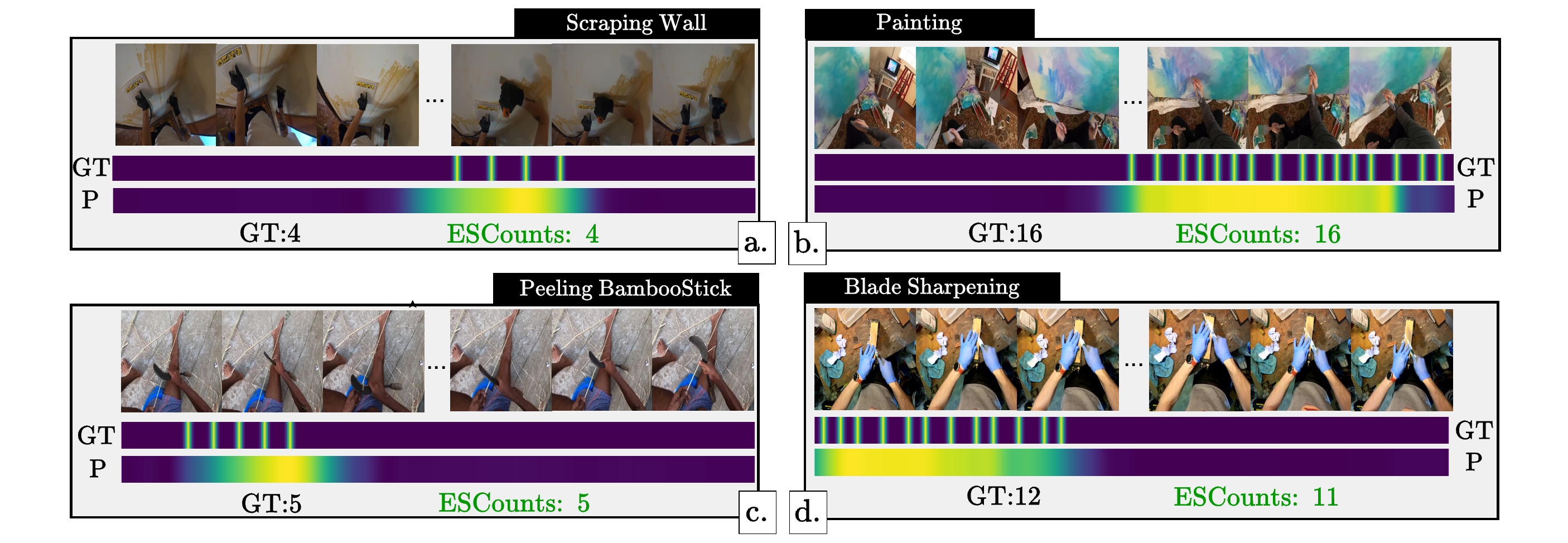}
    \caption{Qualitative results of ESCounts on OVR-Ego4D. For the selected videos, we show both ground truth (GT) and predicted (P) density maps along with the counts. Note that for OVR-Ego4D, we do not have temporal annotations for individual repetitions. Therefore similar to Countix, we show pseudo-labels as the GT density maps.}
    \label{fig:qualitative_ovrego4d}
\end{minipage}
\end{figure}

We evaluate ESCounts on OVR-Ego4D in~\cref{tab:ovrego4d}. Compared to the vision-language-based OVRCounter, \cite{dwibedi2024ovr} ESCounts improves OBZ, OBO, and MAE, with only visual inputs, \textit{without any language input in training or inference}, showing ESCounts' effectiveness for the domain of egocentric counting. We also add some qualitative results in \cref{fig:qualitative_ovrego4d}. Similar to results on other datasets, ESCounts predicts accurate counts a over diverse range of counts. 
The peaks of individual repetitions are not as clear, due to the pseudo-labels, but ESCounts correctly finds the OBO counts in each case.

\end{document}